\documentclass{article}
\PassOptionsToPackage{table}{xcolor}

\usepackage{microtype}
\usepackage{graphicx}
\usepackage{subcaption}
\usepackage{booktabs} 

\usepackage{hyperref}

 \usepackage[preprint]{icml2026}

\usepackage{amsmath}
\usepackage{amssymb}
\usepackage{mathtools}
\usepackage{amsthm}

\usepackage{amsmath, amssymb, amsfonts}
\usepackage{bbm}          

\usepackage{algorithm}
\usepackage{algorithmic}

\usepackage{enumitem}     

\usepackage{multirow}

\usepackage{url}

\usepackage{xcolor}       

\usepackage{colortbl}
\usepackage{adjustbox}
\usepackage{siunitx}

\usepackage{makecell}

\usepackage[capitalize,noabbrev]{cleveref}

\theoremstyle{plain}

\theoremstyle{definition}

\theoremstyle{remark}

\definecolor{posgreen}{RGB}{0,120,0}
\definecolor{negred}{RGB}{180,0,0}
\newcommand{\pos}[1]{\textcolor{posgreen}{#1}}
\newcommand{\negval}[1]{\textcolor{negred}{#1}}

\definecolor{PosGreen}{RGB}{0,130,0}
\definecolor{NegRed}{RGB}{220,30,30}

\usepackage[textsize=tiny]{todonotes}

\icmltitlerunning{Recalibration with Reasoning Priors is all you need for reliable Time-Series Anomaly Detection}

\begin{document}

\twocolumn[
  \icmltitle{Recalibration with Reasoning Priors is all you need for reliable Time-Series Anomaly Detection}




\begin{icmlauthorlist}
    \icmlauthor{Ashish Mishra}{hpe}
    \icmlauthor{Tarun Kumar}{hpe}
    \icmlauthor{Satish Kumar Mopur}{hpe}
    \icmlauthor{Suparna Bhattacharya}{hpe}
    \icmlauthor{Martin Foltin}{hpe}
\end{icmlauthorlist}


\icmlaffiliation{hpe}{
    Hewlett Packard Enterprise,
    Bangalore, India
}


\icmlcorrespondingauthor{Ashish Mishra}{mishra@cse.iitm.ac.in}
\icmlcorrespondingauthor{Tarun Kumar}{}
\icmlcorrespondingauthor{Satish Kumar Mopur}{}
\icmlcorrespondingauthor{Suparna Bhattacharya}{}
\icmlcorrespondingauthor{Martin Foltin}{}

  \icmlkeywords{Machine Learning, ICML}

  \vskip 0.3in
]



\printAffiliationsAndNotice{}  

\begin{abstract}
Time-series anomaly detection is increasingly evaluated under event- and range-aware protocols, yet most detectors---from classical statistical methods to deep models and recent time-series foundation models (TSFMs)---still share a common failure mode: their anomaly scores are often miscalibrated, sensitive to shifts (e.g., missingness, regime changes), and provide little actionable explanation. As a result, detected peaks may not align with true onsets, long events fragment into spikes, and operators lack root-cause cues.

We introduce \textbf{Reasoning-Augmented Anomaly Scoring} (\textsc{RAAS}), a lightweight plug-in layer that \emph{calibrates} the output of any base detector via structured reasoning.
\textsc{RAAS} extracts compact evidence signals (change-points, derivative spikes, variance/slope jumps, spectral drift), generates a small set of mechanistic hypotheses (shift, spike, variance burst, trend change, seasonal drift, missingness burst), and uses a compact LLM-based reranker to score hypothesis--evidence consistency.
The top hypotheses induce an interpretable reasoning score curve which is fused with the base score, yielding event-consistent alarms and explicit explanations (type, onset, root variables, evidence IDs).

On \textsc{TSB-AD} in both multivariate and univariate settings, \textsc{RAAS} improves VUS-PR across diverse detector families, including zero-shot TSFMs, with an average gain of \textbf{+31.9\%} in multivariate VUS-PR (0.177$\rightarrow$0.234) while adding minimal overhead.
These results indicate that \textbf{structured reasoning is a missing layer between raw anomaly scores and robust, explainable anomaly decisions}.

\end{abstract}

\begin{figure*}[t]
    \centering
    \includegraphics[width=\textwidth]{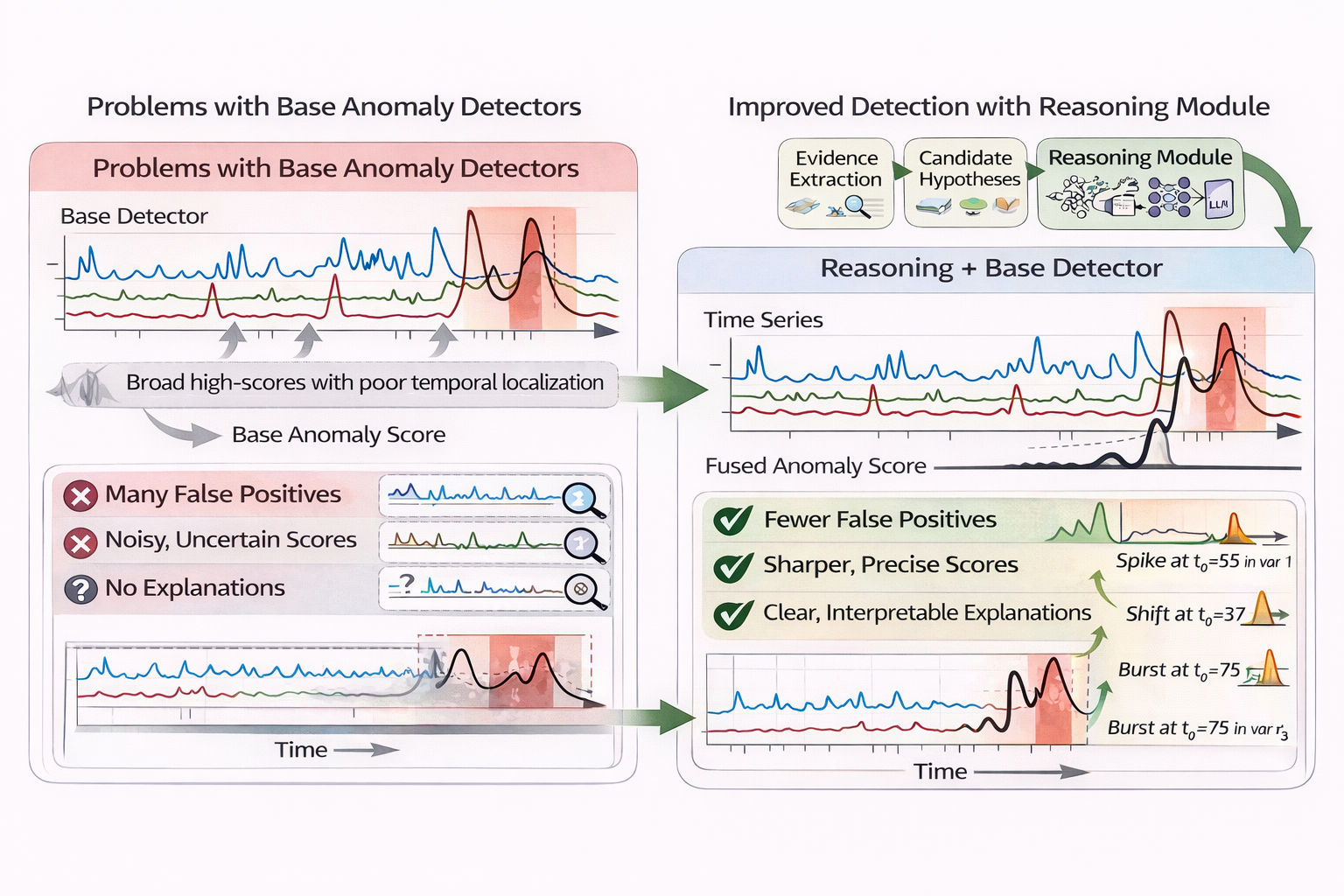}
    \caption{\textbf{Motivation.} Base anomaly detectors often produce noisy or broad anomaly scores with poor temporal localization and no explanations. Our reasoning module converts time-series evidence into human-interpretable hypotheses (e.g., spike, shift, variance burst, trend change), re-ranks them via a lightweight LLM, and fuses the resulting reasoning score with the detector score—reducing false positives and sharpening anomaly localization while providing clear explanations.}
    \label{fig:motivation_reasoning}
\end{figure*}

\begin{figure*}[t]
    \centering
    \includegraphics[width=0.95\textwidth]{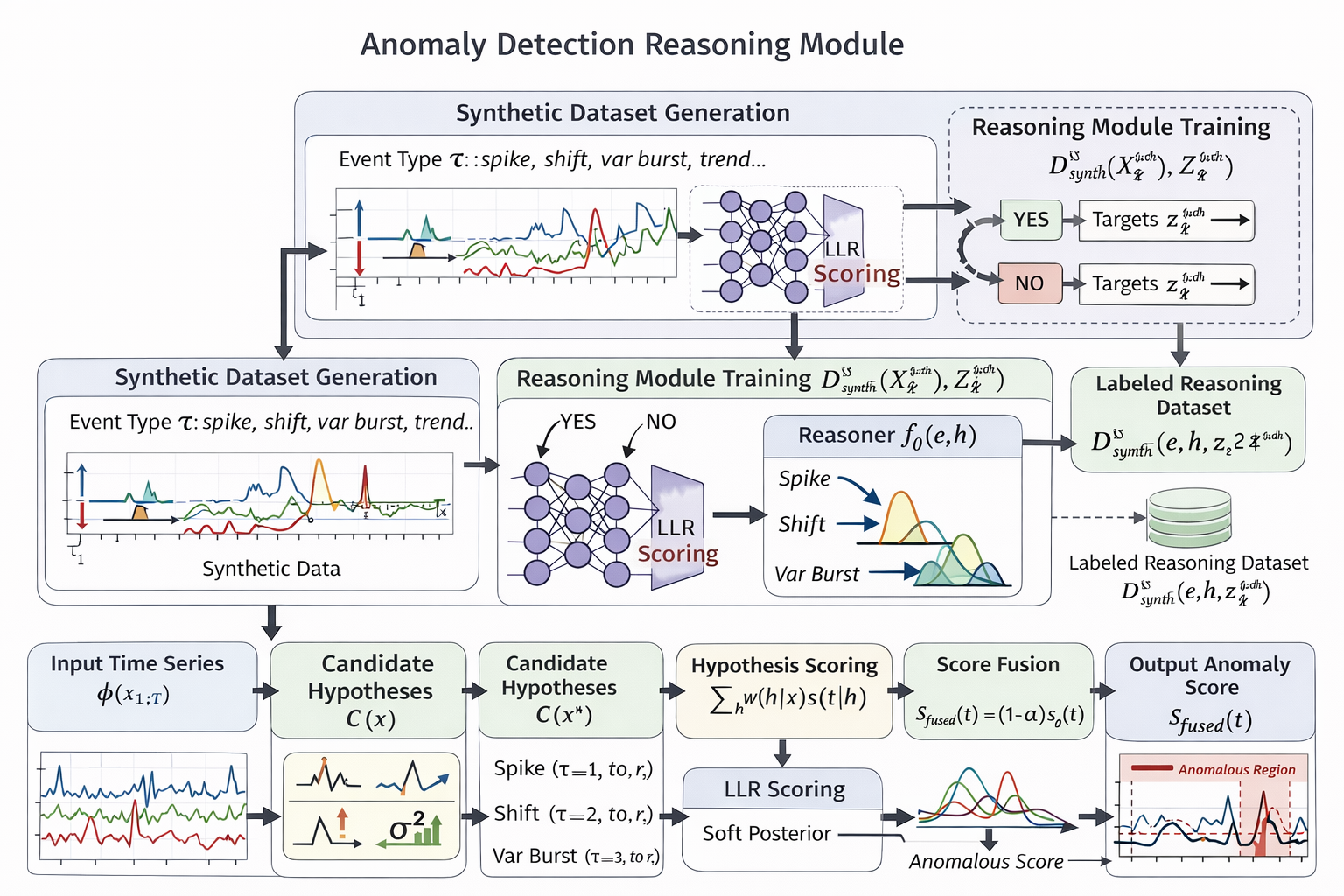}
    \caption{Overall Architecture of our RAAS Approach.  }
    \label{fig:reasoning_approach}
\end{figure*}

\section{Introduction}
Time-series anomaly detection (TSAD) is a core primitive for monitoring and decision-making in cloud services, industrial systems, finance, and healthcare.
In real deployments, an anomaly detector is judged not only by ranking anomalous points, but by whether it produces (i) \emph{reliable scores under distribution shift}, (ii) \emph{event- and range-consistent alarms} (anomalies are intervals with onsets/offsets, not isolated spikes), and (iii) \emph{actionable explanations}, such as the anomaly type and which variables/sensors are responsible.
Despite strong progress, these requirements are rarely satisfied simultaneously.
\vspace{-0.3cm}
\paragraph{Why stronger detectors are not enough.}
The community has steadily advanced from classical unsupervised baselines (e.g., density/distance and subspace methods) to deep temporal architectures and, more recently, time-series foundation models pretrained at scale.
Yet, across detector families we observe a recurring failure mode: the output anomaly scores are often \emph{miscalibrated} and \emph{semantically ambiguous}.
Score peaks may drift away from true onsets, multi-step events can fragment into scattered spikes, and scores can be overly sensitive to missing values or benign regime changes.
These issues become particularly visible under modern evaluation protocols that emphasize event-, range-, and affiliation-aware semantics (e.g., TSB-AD-style evaluation and VUS metrics) \cite{elephant}.
In short, TSAD is not just ``detect large deviations''; it requires aligning score dynamics with the underlying abnormal mechanism.
\vspace{-0.3cm}
\paragraph{Key observation: detectors lack a reasoning layer.}
Although detectors differ in architecture and training, their outputs exhibit similar ambiguities.
For example, a sharp spike and a gradual trend change can receive similarly high scores despite having different temporal signatures and operational implications; in multivariate streams, base scores typically provide little insight into which variables triggered the event.
This suggests that the missing component is not another detector, but a lightweight \emph{reasoning layer} that can interpret evidence patterns, select plausible anomaly mechanisms, and recalibrate scores into event-consistent alarms.
\vspace{-0.3cm}
\paragraph{Reasoning-Augmented Anomaly Scoring (RAAS).}
We propose \textbf{Reasoning-Augmented Anomaly Scoring} (\textsc{RAAS}), a plug-in framework that upgrades \emph{any} anomaly detector with a learned reasoning module.
Given a time series, \textsc{RAAS} (i) extracts a compact set of structured evidence signals (e.g., change-point strength, first-difference spikes, rolling variance jumps, slope changes, and spectral drift), (ii) generates a small hypothesis set over common anomaly mechanisms (shift, spike, variance burst, trend change, seasonal drift, missingness burst), and (iii) uses a compact language-model reranker to judge hypothesis--evidence consistency via a robust YES/NO margin score \cite{qwen2}.
The selected hypotheses are then converted into a smooth, interpretable \emph{reasoning score curve} (via type-specific kernels centered at candidate onsets) and fused with the base detector score.
\textsc{RAAS} produces both an improved anomaly score and an explanation tuple (type, onset time, root variables, evidence IDs).
\vspace{-0.25cm}
\paragraph{Why RAAS helps:}
\textsc{RAAS} provides three practical benefits.
First, it is \emph{detector-agnostic}: it can be applied uniformly to statistical baselines, deep detectors, and zero-shot foundation models without retraining the base detector.
Second, it is \emph{event-aligned}: hypothesis-induced score templates encourage coherent peaks around plausible onsets, improving range- and event-aware evaluation.
Third, it is \emph{explainable}: the output explicitly reports anomaly type, onset time, and (for multivariate series) responsible variables.
\vspace{-0.25cm}
\paragraph{Results highlight.}
On both multivariate and univariate TSAD benchmarks, \textsc{RAAS} consistently improves VUS-PR across a broad set of detectors, including statistical, deep, and zero-shot FM-based anomaly scoring.
For example, on multivariate datasets, we observe an average VUS-PR gain of \textbf{+31.9\%} across all detectors (Table~\ref{tab:vuspr_reasoning}), with particularly large improvements on challenging settings where base scores are fragmented or poorly calibrated.

\paragraph{Contributions.}
\vspace{-0.25cm}
\begin{itemize}[leftmargin=*,itemsep=2pt]
    \item We introduce \textsc{RAAS}, a detector-agnostic reasoning layer that calibrates anomaly scores using structured evidence, hypothesis-level consistency scoring, and score fusion.
    \item We propose a hypothesis-driven, interpretable scoring mechanism that outputs anomaly type, onset time, and root variables with traceable evidence IDs.
    \item We demonstrate consistent improvements across univariate and multivariate benchmarks and across detector families (statistical, deep, and zero-shot foundation models), emphasizing robust event-/range-aware evaluation (VUS-PR) \cite{elephant}.
\end{itemize}


\section{Related Work}
\label{sec:related}

\paragraph{Statistical anomaly detection.}
Classical time-series anomaly detection relies on distributional or geometric cues such as density, distance, and subspace residuals.
Representative approaches include Isolation Forest \cite{iforest}, Local Outlier Factor\cite{lof}, kNN-based outlier scoring \cite{knn}, histogram-based detectors (e.g., HBOS)\cite{hbos}, and PCA/subspace methods \cite{pca}, as well as robust change-point and spectral/seasonality tests.
These methods are typically lightweight and widely used, but they often yield scores that are hard to interpret and can be brittle under mixed anomaly mechanisms.

Beyond detector design, several works \cite{msl, siffer2017anomaly, tatbul2018precision} have studied score calibration and post-processing for anomaly detection, including temperature scaling and probabilistic calibration methods developed for neural classifiers, as well as dynamic thresholding and extreme-value-theory-based score normalization for streaming anomalies. These approaches primarily focus on mapping scores to calibrated probabilities or adaptive thresholds. In contrast, our approach performs mechanism-aware recalibration by reasoning over structured temporal evidence and anomaly hypotheses, producing event-consistent score shapes and explanations rather than only calibrated thresholds.

\paragraph{Deep learning anomaly detection.}
Deep detectors model temporal dynamics and reconstruction/forecasting errors via recurrent networks, autoencoders, variational models, GAN-based methods, and attention/Transformer architectures.
Popular examples include LSTM/GRU autoencoder variants\cite{lstmad}, USAD-style dual autoencoders\cite{usad}, VAE-based methods (e.g., OmniAnomaly)\cite{omnianomaly}, Transformer detectors (e.g., AnomalyTransformer)\cite{anomaly_trans}, and sequence-to-sequence/Transformer reconstructions (e.g., TranAD)\cite{Tranad}, along with diffusion- or self-supervised representations.
While often strong, these methods can still struggle with event consistency (fragmented peaks, threshold sensitivity) and typically provide limited mechanistic explanations beyond saliency or reconstruction error.

\paragraph{Explainable / Root-Cause Time Series Anomaly Detection}
Explainable and root-cause-aware anomaly detection has been explored through graph-based \cite{liu2021root,deng2021graph} and attribution-driven models \cite{weng2018root}, including graph neural network detectors \cite{zhao2020multivariate} and structured probabilistic models \cite{zong2018deep} that localize responsible variables or services. Representative approaches include graph deviation networks, attention-based multivariate detectors, and deep mixture models that jointly model reconstruction and density. These methods embed attribution within detector training. In contrast, our method is detector-agnostic and performs post-hoc mechanistic reasoning over extracted evidence, enabling root-variable attribution and anomaly-type explanations without retraining the base model.

\paragraph{Foundation-model-based anomaly detection.}
Recent work explores time-series foundation models (FMs) pretrained on large-scale corpora and adapted in a zero-shot or light prompting/finetuning manner for downstream tasks, including anomaly detection.
Representative models include TimesFM \cite{timesfm}, MOMENT\cite{moment}, Chronos(-Bolt)\cite{chronos}, Time-MoE and other mixture-of-experts time-series FMs, as well as large generative forecasters used as anomaly scorers via likelihood/residual signals.
Although promising for transfer, FM-based anomaly scores can be poorly calibrated across domains and still lack structured interpretability (type/onset/root causes)\cite{timercb}, \cite{ModernTCN}, \cite{perimid-former}.

\paragraph{Positioning.}
Unlike prior detectors, our method is \emph{detector-agnostic}: it augments a base score with a lightweight reasoning that converts structured evidence into hypothesis-level consistency judgments and a smooth reasoning score curve, yielding improved VUS-PR \emph{and} explicit anomaly explanations (type, onset, root variables).

\section{Method}
\label{sec:method}

\subsection{Problem Setup}
Let $X \in \mathbb{R}^{T \times D}$ denote a multivariate time series with length $T$ and $D$ variables.
We assume access to a \emph{base anomaly detector} $\mathcal{A}$ (statistical, deep, or foundation-model based),
which outputs a real-valued anomaly score sequence:
\begin{equation}
s_{\text{base}}(t) = \mathcal{A}(X)_t, \quad t=1,\dots,T.
\end{equation}
Higher values indicate higher anomaly likelihood.

Our goal is to enhance $s_{\text{base}}$ using a lightweight \emph{reasoning module}
that (i) extracts compact evidence from $X$, (ii) proposes plausible anomaly hypotheses,
(iii) uses a trained language-model reranker to judge hypothesis consistency, and
(iv) converts the best explanations into a complementary anomaly score curve that is fused with $s_{\text{base}}$.

\vspace{2pt}
\noindent\textbf{Outputs.}
Beyond an improved score $s_{\text{fused}}(t)$, our module yields an interpretable explanation:
\[
\hat{h} = (\hat{\tau}, \hat{t}_0, \widehat{\mathcal{R}}, \widehat{\mathcal{E}}),
\]
where $\hat{\tau}$ is anomaly type (e.g., \textsc{shift}, \textsc{spike}), $\hat{t}_0$ is onset time,
$\widehat{\mathcal{R}}$ is the set of root variables, and $\widehat{\mathcal{E}}$ are evidence IDs supporting the decision.

\subsection{Reasoning Module Overview}
Our reasoning module is an \emph{LLM-based discriminative reranker} operating over structured facts.
It follows the pipeline:
\begin{equation}
\begin{aligned}
X &\xrightarrow{\;\mathcal{E}\;} \mathcal{F}
   \xrightarrow{\;\mathcal{G}\;} \mathcal{H} \\
&\xrightarrow{\;\mathcal{R}_\theta\;} \{w_m\}_{m=1}^M \\
&\xrightarrow{\;\mathcal{S}\;} s_{\text{reason}}(t)
   \xrightarrow{\;\text{Fuse}\;} s_{\text{fused}}(t),
\end{aligned}
\end{equation}
where:
\begin{itemize}[leftmargin=*]
\item $\mathcal{E}$ extracts \textbf{evidence facts} $\mathcal{F}$ from $X$ (e.g., change-points, spikes, variance jumps).
\item $\mathcal{G}$ generates a finite \textbf{candidate hypothesis set} $\mathcal{H}=\{h_m\}_{m=1}^M$.
\item $\mathcal{R}_\theta$ is a \textbf{trained language model} that scores each hypothesis via a YES/NO decision.
\item $\mathcal{S}$ converts hypothesis weights into a smooth \textbf{reasoning score curve} $s_{\text{reason}}(t)$.
\end{itemize}

\subsection{Evidence Extraction with IDs}
We compress the time series into a small set of interpretable facts $\mathcal{F}$.
Each fact is associated with an evidence ID, enabling traceability in explanations.
In practice we compute multiple statistics over sliding windows around each time $t$.

\paragraph{Univariate evidence.}
For $D=1$, we compute:
\begin{itemize}[leftmargin=*]
\item \textbf{Change-point score} (\texttt{CPk}): mean jump between left/right windows,
\begin{equation}
\mathrm{CP}(t) = \left|\mu(x_{t-w:t-1}) - \mu(x_{t:t+w-1})\right|.
\end{equation}
We retain top-$K$ separated peaks $\{t_k\}$ and record facts \texttt{CP1..CPK}.
\item \textbf{Spike evidence} (\texttt{Diff1}): maximal absolute first difference,
\begin{equation}
\mathrm{Diff}(t) = |x_t - x_{t-1}|, \quad t_{\text{spk}} = \arg\max_t \mathrm{Diff}(t).
\end{equation}
\item \textbf{Variance jump} (\texttt{Var1}): rolling variance discontinuity,
\begin{equation}
\mathrm{VarJ}(t) = \left|\mathrm{Var}(x_{t-w:t-1}) - \mathrm{Var}(x_{t:t+w-1})\right|.
\end{equation}
\item \textbf{Local slope change} (\texttt{Slope1}): difference of local linear trends,
\begin{equation}
\mathrm{SlopeJ}(t) = \left|a(x_{t-w:t-1}) - a(x_{t:t+w-1})\right|,
\end{equation}
where $a(\cdot)$ is the least-squares slope.
\item \textbf{Seasonality drift} (\texttt{Band1}): energy change in a frequency band using Welch PSD.
\end{itemize}

\paragraph{Multivariate evidence.}
For $D>1$, we compute each statistic per variable $d$ and keep top contributing variables.
For example, a multivariate change-point evidence at time $t$ is:
\begin{equation}
\mathrm{CP}_d(t) = \left|\mu(X_{t-w:t-1,d}) - \mu(X_{t:t+w-1,d})\right|.
\end{equation}
We define a root-score vector $r(t)\in\mathbb{R}^D$ by normalizing these per-variable evidences,
and select root candidates $\mathcal{R}(t)$ as the top-$R$ variables with highest $r_d(t)$.
This yields evidence IDs of the form \texttt{CPk(var=d, t=...)} and supports root attribution.

\paragraph{Fact string.}
We serialize facts into a compact text sequence:
\[
\mathrm{FactsText}(\mathcal{F}) = \texttt{ID=value | ID=value | ...},
\]
which serves as the input context for the language reranker.

\subsection{Hypothesis Space}
A hypothesis $h$ describes an interpretable anomaly explanation:
\begin{equation}
h=(\tau, t_0, \mathcal{R}, \mathcal{E}),
\end{equation}
where:
$\tau$ is a type, $t_0$ is onset time, $\mathcal{R}\subseteq\{1,\dots,D\}$ is root set, and
$\mathcal{E}$ is the list of evidence IDs used.

We use a small set of anomaly types that cover common benchmark patterns:
\vspace{-0.5cm}
\begin{multline}
\tau \in \{\textsc{shift}, \textsc{spike},\\
\textsc{variance\_burst}, \textsc{trend\_change},\\
\textsc{seasonal\_drift}, \textsc{missing\_burst}\}.
\end{multline}

Candidate onsets $t_0$ are proposed using evidence peaks (e.g., top change-points, spike time, variance-jump time),
and root variables are proposed via the top-$R$ multivariate evidence contributors.

This yields a small finite candidate set $\mathcal{H}=\{h_m\}_{m=1}^M$, typically $M\le 8$,
enabling efficient reranking.

\subsection{LLM Reranker as a YES/NO Consistency Judge}
We train a small causal language model $\mathcal{R}_\theta$ (e.g., GPT-Neo-125M) to act as a
\emph{binary consistency classifier}:
\[
\mathcal{R}_\theta(\mathrm{FactsText}(\mathcal{F}), h) \rightarrow \{\texttt{YES},\texttt{NO}\}.
\]

\paragraph{Prompt format.}
Each candidate is converted to a prompt:
\begin{quote}
\small
\texttt{You are a time-series anomaly reasoning module.} \\
\texttt{Given FACTS, judge whether the HYPOTHESIS is consistent.} \\
\texttt{FACTS: ...} \\
\texttt{HYPOTHESIS: Type=...; t0=...; root=...} \\
\texttt{USED\_EVIDENCE\_IDS: [ ... ]} \\
\texttt{Answer ONLY one token: YES or NO.}
\end{quote}

\paragraph{Log-probability scoring.}
At inference, we score a hypothesis by the log-likelihood margin:
\begin{equation}
\mathrm{Score}_\theta(h) =
\log p_\theta(\texttt{YES}\mid \mathcal{F},h) - \log p_\theta(\texttt{NO}\mid \mathcal{F},h).
\label{eq:margin}
\end{equation}
This requires only a single forward pass per completion token and is robust to generation formatting.

\paragraph{Soft selection.}
We convert scores into mixture weights over top-$M'$ hypotheses:
\begin{equation}
w_m = \frac{\exp(\mathrm{Score}_\theta(h_m)/\beta)}
{\sum_{j=1}^{M'}\exp(\mathrm{Score}_\theta(h_j)/\beta)},
\end{equation}
where $\beta$ is a temperature (default $\beta=1$).
We retain the single best explanation $\hat{h}=\arg\max_h \mathrm{Score}_\theta(h)$,
and use $\{w_m\}$ to build a smooth score curve.

\subsection{From Hypotheses to a Reasoning Score Curve}
Each hypothesis $h_m=(\tau,t_0,\mathcal{R},\mathcal{E})$ is mapped to a deterministic, interpretable score curve
$s_\tau(t;t_0)$ via type-specific kernels.

\paragraph{Type-specific kernels.}
We use smooth localized kernels centered at $t_0$:
\begin{align}
s_{\textsc{shift}}(t;t_0)
 &= \Delta\mu(t_0)\cdot \notag\\
 &\quad \exp\!\left(-\frac{(t-t_0)^2}{2\sigma^2}\right),\\
s_{\textsc{spike}}(t;t_0)
 &= \max|x_t-x_{t-1}|\cdot \notag\\
 &\quad \exp\!\left(-\frac{(t-t_0)^2}{2\sigma_{\text{spk}}^2}\right),\\
s_{\textsc{variance\_burst}}(t;t_0)
 &= \Delta\mathrm{Var}(t_0)\cdot \notag\\
 &\quad \exp\!\left(-\frac{(t-t_0)^2}{2\sigma_{\text{var}}^2}\right),\\
s_{\textsc{trend\_change}}(t;t_0)
 &= \Delta a(t_0)\cdot \notag\\
 &\quad \exp\!\left(-\frac{(t-t_0)^2}{2\sigma_{\text{tr}}^2}\right).
\end{align}
and similarly for \textsc{seasonal\_drift} (band energy change) and \textsc{missing\_burst} (NaN indicator).

\paragraph{Mixture score.}
We combine kernels using reranker weights:
\begin{equation}
s_{\text{reason}}(t) = \sum_{m=1}^{M'} w_m \cdot s_{\tau_m}(t; t_{0,m}).
\label{eq:reason_curve}
\end{equation}
We then normalize $s_{\text{reason}} \in [0,1]$. This provides a detector-agnostic anomaly curve grounded in interpretable evidence.

\subsection{Fusion with Any Base Detector}
We fuse the base detector score with the reasoning curve:
\begin{equation}
s_{\text{fused}}(t) = \mathrm{Norm}\Big((1-\alpha)s_{\text{base}}(t) + \alpha s_{\text{reason}}(t)\Big),
\label{eq:fuse}
\end{equation}
where $\alpha\in[0,1]$ controls the reasoning contribution.
In experiments we set $\alpha\in\{0.1,0.2,0.3\}$ and see stable gains.

\paragraph{Theoretical justifications:}
We have provided the theoretical justification about our datasets generation and hypothesis preparation in Appendix \ref{sec:synthetic_completeness} and \ref{sec:theory}.

\section{Experiments and Results}


\subsubsection{Dataset and Baseline}
In our experiments, we consider four widely used multivariate anomaly detection benchmarks (in their redesigned public versions as in\cite{elephant}): SMD \cite{smd}, MSL \cite{msl}, PSM \cite{psm}, and SMAP~\cite{msl}. We additionally evaluate the univariate setting on challenging and commonly used datasets, including MGAB~\cite{mgab}, NAB~\cite{nab}, and NEK~\cite{nek}.

To fine-tune the reasoner, we generated synthetic data (detailed process in Appendix \ref{subsec:appendix:synthetic}). We apply our reasoning module across three families of base detectors: (i) statistical methods—IForest~\cite{iforest}, LOF~\cite{lof}, HBOS~\cite{hbos}, KNN~\cite{knn}, and PCA~\cite{pca}; (ii) deep learning methods—AnomalyTransformer~\cite{anomaly_trans}, TranAD~\cite{Tranad}, USAD~\cite{usad}, OmniAnomaly~\cite{omnianomaly}, and LSTMAD~\cite{lstmad}; and (iii) zero-shot time-series foundation models—TimesFM~\cite{timesfm}, MOMENT~\cite{moment}, Chronos~\cite{chronos}, and Times-MoE~\cite{timemoe}. 


\subsubsection{Evaluation Metrics}
Prior work has noted that commonly used metrics for multivariate time-series anomaly detection—such as F1, AUC-PR, AUC-ROC, and Affiliation-F1—can suffer from evaluation pitfalls, whereas VUS-PR provides a more robust, accurate, and fair assessment~\cite{elephant}. Accordingly, we adopt VUS-PR as our primary metric. For completeness, we also report the other metrics, and provide additional results (including hyperparameter analyses and visualizations) in the Appendix.

We use Qwen2-0.5B-Instruct\cite{qwen2} as LLM Reasoner in our approach.

\subsection{Results}
Table~\ref{tab:vuspr_reasoning} shows that augmenting a base detector with our reasoning module consistently improves VUS-PR across \emph{statistical}, \emph{deep}, and \emph{zero-shot foundation-model} (FM) detectors, without retraining the detector.
Overall, the average VUS-PR increases from \textbf{0.177} to \textbf{0.234} (\textbf{+31.9\%}).
Improvements are strongest on datasets where raw detector scores are less stable or less semantically grounded, e.g.,
\textbf{SAMP} improves by \textbf{+56.6\%} on average (0.115$\rightarrow$0.180), and \textbf{SMD}/\textbf{MSL} improve by \textbf{+40.8\%}/\textbf{+38.1\%}.
Notably, the module remains effective even for zero-shot FMs (e.g., \textbf{Time-MoE} +39.6\%, \textbf{Chronos} +37.8\%, \textbf{Moment} +34.5\%, \textbf{TimesFM} +27.7\% average),
demonstrating that reasoning provides a \emph{detector-agnostic} mechanism to refine anomaly scores by enforcing
type/onset/root consistency and producing a smoother, evidence-aligned scoring curve. For additional results, please refer to Appendix \ref{sec:appendix:detailed-results}.

\begin{table*}[t]
\centering
\scriptsize
\setlength{\tabcolsep}{1.8pt}
\renewcommand{\arraystretch}{1.02}
\caption{\textbf{VUS-PR comparison with/without the reasoning module} on multivariate time-series anomaly detection.
``Base'' is the detector alone and ``RAAS'' augments it with our reasoning module.
$\Delta\%$ is relative improvement.
FM detectors are \textbf{zero-shot}.}
\label{tab:vuspr_reasoning}

\resizebox{\textwidth}{!}{%
\begin{tabular}{l|rrr|rrr|rrr|rrr|rrr}
\toprule
\multirow{2}{*}{\textbf{Model}} &
\multicolumn{3}{c|}{\textbf{SMD}} &
\multicolumn{3}{c|}{\textbf{MSL}} &
\multicolumn{3}{c|}{\textbf{PSM}} &
\multicolumn{3}{c|}{\textbf{SAMP}} &
\multicolumn{3}{c}{\textbf{Avg (4 datasets)}} \\
\cmidrule(lr){2-4}\cmidrule(lr){5-7}\cmidrule(lr){8-10}\cmidrule(lr){11-13}\cmidrule(lr){14-16}
& \textbf{Base} & \textbf{RAAS} & \textbf{$\Delta$\%}
& \textbf{Base} & \textbf{RAAS} & \textbf{$\Delta$\%}
& \textbf{Base} & \textbf{RAAS} & \textbf{$\Delta$\%}
& \textbf{Base} & \textbf{RAAS} & \textbf{$\Delta$\%}
& \textbf{Base} & \textbf{RAAS} & \textbf{$\Delta$\%} \\
\midrule

\multicolumn{16}{l}{\textbf{Statistical (5)}} \\
HBOS    & 0.254 & 0.288 & \pos{+13.4\%} & 0.208 & 0.292 & \pos{+40.4\%} & 0.158 & 0.158 & \pos{+0.0\%}  & 0.090 & 0.195 & \pos{+116.7\%} & 0.177 & 0.233 & \pos{+31.4\%} \\
IForest & 0.213 & 0.241 & \pos{+13.1\%} & 0.441 & 0.509 & \pos{+15.4\%} & 0.162 & 0.162 & \pos{+0.0\%}  & 0.370 & 0.412 & \pos{+11.4\%}  & 0.296 & 0.331 & \pos{+11.6\%} \\
KNN     & 0.281 & 0.372 & \pos{+32.4\%} & 0.169 & 0.296 & \pos{+75.1\%} & 0.116 & 0.117 & \pos{+0.9\%}  & 0.073 & 0.212 & \pos{+190.4\%} & 0.160 & 0.249 & \pos{+55.8\%} \\
LOF     & 0.126 & 0.269 & \pos{+113.5\%}& 0.140 & 0.313 & \pos{+123.6\%}& 0.142 & 0.142 & \pos{+0.0\%}  & 0.076 & 0.218 & \pos{+186.8\%} & 0.121 & 0.236 & \pos{+95.8\%} \\
PCA     & 0.533 & 0.554 & \pos{+3.9\%}  & 0.381 & 0.467 & \pos{+22.6\%} & 0.176 & 0.176 & \pos{+0.0\%}  & 0.309 & 0.350 & \pos{+13.3\%}  & 0.350 & 0.387 & \pos{+10.5\%} \\

\midrule
\multicolumn{16}{l}{\textbf{Deep Learning (5)}} \\
AnomalyTransformer & 0.053 & 0.215 & \pos{+305.7\%} & 0.097 & 0.401 & \pos{+313.4\%} & 0.123 & 0.123 & \pos{+0.0\%}  & 0.036 & 0.250 & \pos{+594.4\%} & 0.077 & 0.247 & \pos{+221.1\%} \\
LSTMAD             & 0.159 & 0.234 & \pos{+47.2\%}  & 0.254 & 0.281 & \pos{+10.6\%}  & 0.347 & 0.349 & \pos{+0.6\%}  & 0.071 & 0.099 & \pos{+39.4\%}  & 0.208 & 0.241 & \pos{+16.0\%} \\
OmniAnomaly         & 0.120 & 0.146 & \pos{+21.7\%}  & 0.254 & 0.275 & \pos{+8.3\%}   & 0.261 & 0.270 & \pos{+3.4\%}  & 0.067 & 0.081 & \pos{+20.9\%}  & 0.176 & 0.193 & \pos{+9.6\%} \\
TranAD             & 0.162 & 0.256 & \pos{+58.0\%}  & 0.234 & 0.239 & \pos{+2.1\%}   & 0.327 & 0.330 & \pos{+0.9\%}  & 0.065 & 0.092 & \pos{+41.5\%}  & 0.197 & 0.229 & \pos{+16.1\%} \\
USAD               & 0.131 & 0.168 & \pos{+28.2\%}  & 0.254 & 0.342 & \pos{+34.6\%}  & 0.314 & 0.322 & \pos{+2.5\%}  & 0.066 & 0.073 & \pos{+10.6\%}  & 0.191 & 0.226 & \pos{+18.4\%} \\

\midrule
\multicolumn{16}{l}{\textbf{Foundation Models (Zero-shot) (4)}} \\
Moment (ZS)   & 0.054 & 0.091 & \pos{+68.5\%} & 0.103 & 0.143 & \pos{+38.8\%} & 0.097 & 0.097 & \pos{+0.0\%}  & 0.089 & 0.132 & \pos{+47.8\%} & 0.086 & 0.115 & \pos{+34.5\%} \\
TimesFM (ZS)  & 0.175 & 0.345 & \pos{+97.1\%} & 0.452 & 0.496 & \pos{+9.7\%}  & 0.138 & 0.139 & \pos{+0.7\%}  & 0.170 & 0.215 & \pos{+26.8\%} & 0.234 & 0.299 & \pos{+27.7\%} \\
Chronos (ZS)  & 0.116 & 0.156 & \pos{+34.5\%} & 0.086 & 0.171 & \pos{+99.8\%} & 0.145 & 0.155 & \pos{+6.6\%}  & 0.074 & 0.094 & \pos{+27.0\%} & 0.105 & 0.144 & \pos{+37.8\%} \\
Time-MoE (ZS) & 0.127 & 0.191 & \pos{+50.4\%} & 0.078 & 0.128 & \pos{+62.4\%} & 0.157 & 0.162 & \pos{+3.2\%}  & 0.050 & 0.091 & \pos{+83.8\%} & 0.103 & 0.143 & \pos{+39.6\%} \\

\midrule
\textbf{Avg (All models)} &
\textbf{0.179} & \textbf{0.252} & \pos{\textbf{+40.8\%}} &
\textbf{0.225} & \textbf{0.311} & \pos{\textbf{+38.1\%}} &
\textbf{0.190} & \textbf{0.193} & \pos{\textbf{+1.4\%}}  &
\textbf{0.115} & \textbf{0.180} & \pos{\textbf{+56.6\%}} &
\textbf{0.177} & \textbf{0.234} & \pos{\textbf{+31.9\%}} \\
\bottomrule
\end{tabular}
}
\end{table*}

\begin{figure*}[t]
  \centering
  \setlength{\tabcolsep}{6pt}
  \begin{tabular}{cc}
    \includegraphics[width=0.48\textwidth]{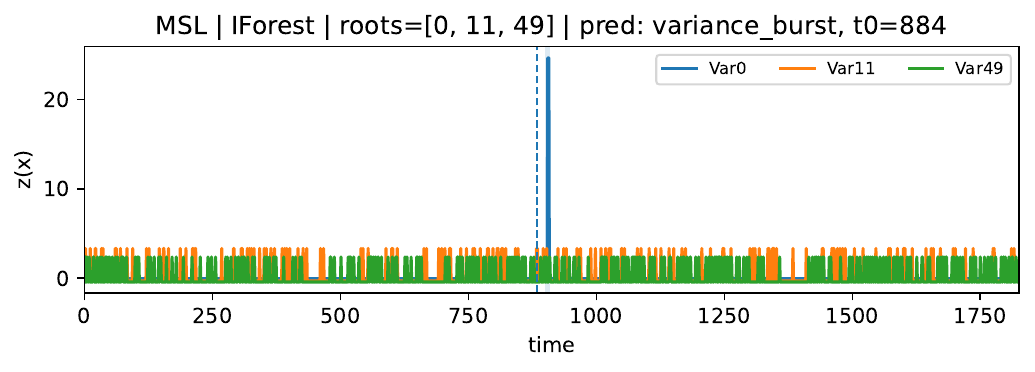} &
    \includegraphics[width=0.48\textwidth]{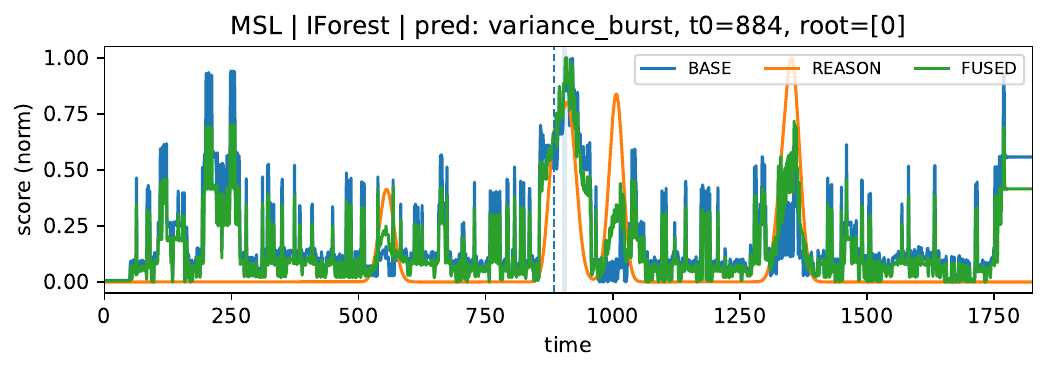} \\
  \end{tabular}
  \caption{\textbf{Example explanation produced by our reasoning-augmented detector (MSL + IForest).}
  \textbf{Left:} predicted anomaly type (\texttt{variance\_burst}), onset $t_0$, and attributed root variables (dominant sensors).
  \textbf{Right:} base score vs reasoning score vs fused score, showing that reasoning sharpens and stabilizes the anomaly peak around the predicted onset.}
  \label{fig:explain_msl_iforest}
\end{figure*}

\paragraph{Qualitative explainability.}
Figure~\ref{fig:explain_msl_iforest} illustrates that our reasoning module not only refines the anomaly score but also produces an interpretable explanation in terms of \emph{anomaly type}, \emph{onset time}, and \emph{root variables}.
In this example on MSL with IForest, the module predicts a \texttt{variance\_burst} with onset around $t_0{=}884$ and attributes the event to a small set of dominant sensors (e.g., roots $[0,11,49]$), providing actionable localization of the anomaly source.
Correspondingly, the reasoning-induced curve forms a smooth, evidence-aligned peak around $t_0$ and the fused score follows this corrected structure, improving event consistency compared to the raw base score. We have included more visualization results in Appendix \ref{sec:appendix:ablation}.

\begin{table*}[t]
\centering
\scriptsize
\setlength{\tabcolsep}{1.45pt}     
\renewcommand{\arraystretch}{0.98} 
\caption{\textbf{Univariate anomaly detection (TSB-AD-U): VUS-PR (higher is better).}
We report \textbf{Base} vs \textbf{RAAS} mean VUS-PR over files for each dataset.
$\Delta\%$ is relative improvement.
\textbf{Avg} summarizes the mean across the 5 datasets, and \textbf{Avg $\Delta\%$} is the relative gain on that mean.}
\label{tab:tsb_adu_vuspr_compact}

\resizebox{\textwidth}{!}{%
\begin{tabular}{ll|rrr|rrr|rrr|rrr|rrr|rrr}
\toprule
\multirow{2}{*}{\textbf{Family}} & \multirow{2}{*}{\textbf{Detector}} &
\multicolumn{3}{c|}{\textbf{IOPS}} &
\multicolumn{3}{c|}{\textbf{MGAB}} &
\multicolumn{3}{c|}{\textbf{NAB}} &
\multicolumn{3}{c|}{\textbf{NEK}} &
\multicolumn{3}{c|}{\textbf{Power}} &
\multicolumn{3}{c}{\textbf{Avg (5 datasets)}} \\
\cmidrule(lr){3-5}\cmidrule(lr){6-8}\cmidrule(lr){9-11}\cmidrule(lr){12-14}\cmidrule(lr){15-17}\cmidrule(lr){18-20}
& &
\textbf{Base} & \textbf{RAAS} & \textbf{$\Delta$\%} &
\textbf{Base} & \textbf{RAAS} & \textbf{$\Delta$\%} &
\textbf{Base} & \textbf{RAAS} & \textbf{$\Delta$\%} &
\textbf{Base} & \textbf{RAAS} & \textbf{$\Delta$\%} &
\textbf{Base} & \textbf{RAAS} & \textbf{$\Delta$\%} &
\textbf{Base} & \textbf{RAAS} & \textbf{$\Delta$\%} \\
\midrule

\multirow{5}{*}{\textbf{Stat}} 
& IForest & 0.133 & 0.146 & \pos{+9.8\%}  & 0.005 & 0.006 & \pos{+20.0\%} & 0.321 & 0.357 & \pos{+11.2\%} & 0.774 & 0.842 & \pos{+8.8\%}  & 0.075 & 0.077 & \pos{+2.7\%}  & 0.2616 & 0.2856 & \pos{+9.2\%}  \\
& LOF     & 0.081 & 0.19  & \pos{+134.6\%}& 0.006 & 0.008 & \pos{+33.3\%} & 0.153 & 0.424 & \pos{+177.1\%}& 0.45  & 0.843 & \pos{+87.3\%} & 0.09  & 0.104 & \pos{+15.6\%} & 0.156  & 0.3138 & \pos{+101.2\%} \\
& HBOS    & 0.03  & 0.153 & \pos{+410.0\%}& 0.006 & 0.009 & \pos{+50.0\%} & 0.126 & 0.555 & \pos{+340.5\%}& 0.168 & 0.718 & \pos{+327.4\%}& 0.09  & 0.101 & \pos{+12.2\%} & 0.084  & 0.3072 & \pos{+265.7\%} \\
& KNN     & 0.03  & 0.153 & \pos{+410.0\%}& 0.006 & 0.009 & \pos{+50.0\%} & 0.126 & 0.555 & \pos{+340.5\%}& 0.168 & 0.718 & \pos{+327.4\%}& 0.09  & 0.101 & \pos{+12.2\%} & 0.084  & 0.3072 & \pos{+265.7\%} \\
& PCA     & 0.222 & 0.24  & \pos{+8.1\%}  & 0.007 & 0.008 & \pos{+14.3\%} & 0.46  & 0.475 & \pos{+3.3\%}  & 0.852 & 0.87  & \pos{+2.1\%}  & 0.076 & 0.079 & \pos{+3.9\%}  & 0.3234 & 0.3344 & \pos{+3.4\%}  \\
\midrule

\multirow{5}{*}{\textbf{DL}}
& AnomalyTransformer & 0.035 & 0.154 & \pos{+340.0\%}& 0.007 & 0.01  & \pos{+42.9\%} & 0.136 & 0.547 & \pos{+302.2\%}& 0.168 & 0.714 & \pos{+325.0\%}& 0.09  & 0.108 & \pos{+20.0\%} & 0.0872 & 0.3066 & \pos{+251.6\%} \\
& TranAD            & 0.039 & 0.071 & \pos{+82.1\%} & 0.007 & 0.008 & \pos{+14.3\%} & 0.272 & 0.382 & \pos{+40.4\%} & 0.653 & 0.753 & \pos{+15.3\%} & 0.088 & 0.086 & \negval{-2.3\%}  & 0.2118 & 0.26   & \pos{+22.8\%} \\
& USAD              & 0.04  & 0.056 & \pos{+40.0\%} & 0.007 & 0.008 & \pos{+14.3\%} & 0.282 & 0.365 & \pos{+29.4\%} & 0.677 & 0.745 & \pos{+10.0\%} & 0.088 & 0.086 & \negval{-2.3\%}  & 0.2188 & 0.252  & \pos{+15.2\%} \\
& OmniAnomaly       & 0.041 & 0.048 & \pos{+17.1\%} & 0.008 & 0.009 & \pos{+12.5\%} & 0.347 & 0.392 & \pos{+13.0\%} & 0.674 & 0.698 & \pos{+3.6\%}  & 0.088 & 0.087 & \negval{-1.1\%}  & 0.2316 & 0.2468 & \pos{+6.6\%}  \\
& LSTMAD            & 0.046 & 0.109 & \pos{+137.0\%}& 0.018 & 0.022 & \pos{+22.2\%} & 0.296 & 0.397 & \pos{+34.1\%} & 0.606 & 0.769 & \pos{+26.9\%} & 0.088 & 0.086 & \negval{-2.3\%}  & 0.2108 & 0.2766 & \pos{+31.2\%} \\
\midrule

\multirow{4}{*}{\textbf{FM}}
& TimesFM     & 0.179 & 0.328  & \pos{+83.2\%} & 0.006  & 0.007 & \pos{+16.7\%} & 0.172  & 0.378  & \pos{+119.8\%} & 0.412 & 0.869 & \pos{+110.9\%} & 0.079 & 0.089 & \pos{+12.7\%} & 0.1696 & 0.3342 & \pos{+97.1\%} \\
& MOMENT (ZS)  & 0.203 & 0.248  & \pos{+22.2\%} & 0.006  & 0.007 & \pos{+16.7\%} & 0.517  & 0.538  & \pos{+4.1\%}   & 0.703 & 0.755 & \pos{+7.4\%}   & 0.073 & 0.075 & \pos{+2.7\%}  & 0.3004 & 0.3246 & \pos{+8.1\%}  \\
& Chronos      & 0.1925 & 0.215 & \pos{+11.7\%} & 0.0067 & 0.0081 & \pos{+20.9\%} & 0.237  & 0.2825 & \pos{+19.2\%}  & 0.325 & 0.3675 & \pos{+13.1\%}  & 0.091 & 0.11   & \pos{+20.9\%} & 0.1704 & 0.1966 & \pos{+15.4\%} \\
& Times-MoE    & 0.1663 & 0.1977 & \pos{+18.9\%} & 0.0055 & 0.0068 & \pos{+23.6\%} & 0.2264 & 0.2578 & \pos{+13.9\%}  & 0.1986 & 0.249 & \pos{+25.4\%}  & 0.093 & 0.1056 & \pos{+13.5\%} & 0.1380 & 0.1634 & \pos{+18.4\%} \\
\midrule

\textbf{Avg (All detectors)} &
& \textbf{0.1027} & \textbf{0.1649} & \textbf{\pos{+60.6\%}} &
  \textbf{0.0072} & \textbf{0.0090} & \textbf{\pos{+24.4\%}} &
  \textbf{0.2622} & \textbf{0.4218} & \textbf{\pos{+60.8\%}} &
  \textbf{0.4878} & \textbf{0.7079} & \textbf{\pos{+45.1\%}} &
  \textbf{0.0856} & \textbf{0.0925} & \textbf{\pos{+8.0\%}}  &
  \textbf{0.1891} & \textbf{0.2792} & \textbf{\pos{+47.6\%}} \\
\bottomrule
\end{tabular}
}
\end{table*}

\section{Ablation Study}

\subsection{Fusion-type ablation.}
Table~\ref{tab:fusion_ablation_smd_wide} evaluates whether improvements come from \emph{meaningful} fusion with reasoning, rather than from adding any extra score.
Across three heterogeneous detectors, \textbf{BASERAAS} yields the strongest and most consistent \textbf{VUS-PR} gains (IForest: $+0.104$, TranAD: $+0.103$, TimesFM: $+0.173$), showing that the reasoner provides complementary information that improves ranking quality.
In contrast, \textbf{BASE+random-score} severely degrades VUS-PR for all detectors, serving as a sanity check that fusion alone does not help and that the benefit is specific to the reasoning signal.
\textbf{REASON-only} is competitive but not sufficient: it does not consistently dominate BASE, indicating that the reasoner is best used as a \emph{regularizer/calibrator} rather than a standalone detector.
Finally, \textbf{BASE+rolling-zscore} provides moderate gains, but remains below \textbf{BASERAAS} on VUS-PR, suggesting that simple statistical heuristics capture only limited anomaly structure compared to structured reasoning.
We prioritize VUS-PR here since it is ranking-aware and less sensitive to thresholding effects than F1-style metrics; overall, the ablation confirms that our improvements arise from \emph{structured reasoning-driven fusion}.

\begin{table*}[t]
\centering
\small
\setlength{\tabcolsep}{3.4pt}
\renewcommand{\arraystretch}{1.15}
\caption{\textbf{Fusion ablation on TSB-AD-M (SMD), compact wide layout.}
We average over 22 SMD multivariate files. Each cell shows mean$\pm$std; values in {\scriptsize( )} are $\Delta$ vs the detector's BASE.}
\label{tab:fusion_ablation_smd_wide}
\resizebox{\textwidth}{!}{%
\begin{tabular}{l|cc|cc|cc}
\toprule
\multirow{2}{*}{\textbf{Variant}} &
\multicolumn{2}{c|}{\textbf{IForest}} &
\multicolumn{2}{c|}{\textbf{TranAD}} &
\multicolumn{2}{c}{\textbf{TimesFM}} \\
\cmidrule(lr){2-3}\cmidrule(lr){4-5}\cmidrule(lr){6-7}
& \textbf{VUS-PR}$\uparrow$ & \textbf{Event-F1}$\uparrow$
& \textbf{VUS-PR}$\uparrow$ & \textbf{Event-F1}$\uparrow$
& \textbf{VUS-PR}$\uparrow$ & \textbf{Event-F1}$\uparrow$ \\
\midrule
BASE
& 0.244$\pm$0.062 & 0.596$\pm$0.156
& 0.206$\pm$0.076 & 0.424$\pm$0.196
& 0.175$\pm$0.089 & 0.579$\pm$0.149 \\
REASON-only
& 0.215$\pm$0.104\,{\scriptsize(-0.029)} & 0.575$\pm$0.167\,{\scriptsize(-0.021)}
& 0.215$\pm$0.104\,{\scriptsize(+0.009)} & 0.575$\pm$0.167\,{\scriptsize(+0.151)}
& 0.215$\pm$0.104\,{\scriptsize(+0.040)} & 0.575$\pm$0.167\,{\scriptsize(-0.004)} \\
\textbf{BASE + REASON (ours)}
& \textbf{0.348$\pm$0.114\,{\scriptsize(+0.104)}} & \textbf{0.648$\pm$0.147\,{\scriptsize(+0.052)}}
& \textbf{0.309$\pm$0.120\,{\scriptsize(+0.103)}} & \textbf{0.582$\pm$0.147\,{\scriptsize(+0.158)}}
& \textbf{0.349$\pm$0.114\,{\scriptsize(+0.173)}} & 0.535$\pm$0.162\,{\scriptsize(-0.044)} \\
BASE + random-score
& 0.073$\pm$0.074\,{\scriptsize(-0.171)} & 0.510$\pm$0.195\,{\scriptsize(-0.086)}
& 0.048$\pm$0.049\,{\scriptsize(-0.157)} & 0.387$\pm$0.213\,{\scriptsize(-0.037)}
& 0.056$\pm$0.064\,{\scriptsize(-0.118)} & 0.394$\pm$0.243\,{\scriptsize(-0.184)} \\
BASE + rolling-zscore
& 0.305$\pm$0.095\,{\scriptsize(+0.060)} & 0.597$\pm$0.168\,{\scriptsize(+0.001)}
& 0.256$\pm$0.092\,{\scriptsize(+0.050)} & 0.465$\pm$0.186\,{\scriptsize(+0.041)}
& 0.299$\pm$0.093\,{\scriptsize(+0.124)} & \textbf{0.602$\pm$0.170\,{\scriptsize(+0.023)}} \\
\bottomrule
\end{tabular}}
\end{table*}

\subsection{Alpha sensitivity.}
We analyze the fusion weight $\alpha$ in our score composition
\begin{equation}
s_{\alpha} \;=\; (1-\alpha)\,s_{\text{base}} \;+\; \alpha\,s_{\text{reason}},
\end{equation}
where $\alpha=0$ reduces to the base detector and $\alpha=1$ uses the reasoning module alone.
Figure~\ref{fig:alpha_sensitivity_vuspr} shows a consistent \emph{unimodal} trend across both datasets (SMD, MSL) and all three detectors (TimesFM, Moment, TranAD):
small $\alpha$ under-utilizes reasoning, mid-range $\alpha$ yields the best VUS-PR, and large $\alpha$ degrades performance as the fused score becomes dominated by reasoning-only signals.
This behavior supports our hypothesis that the base detector and the reasoner are \emph{complementary}---the detector provides fine-grained temporal sensitivity while the reasoner acts as a calibration/consistency prior.
Concretely, TimesFM and Moment peak at $\alpha=0.4$ on both SMD and MSL, while TranAD peaks at $\alpha=0.5$ on SMD (and $\alpha=0.4$ on MSL), indicating that noisier base scores benefit from slightly stronger reasoning contributions.
For larger $\alpha$, performance drops because reasoning-only scoring cannot fully replace detector-specific inductive biases (e.g., it may over-smooth peaks or mis-rank ambiguous events).
Based on this sweep, we adopt a fixed mid-range setting ($\alpha\approx0.4$; $\alpha=0.5$ for TranAD on SMD), which consistently lies near the optimum and is robust to small deviations.

\subsection{Candidate-generation ablation (coverage--cost tradeoff).}
Table~\ref{tab:candgen_ablation_compact} (in Appendix) isolates the \emph{candidate-generation} stage of our reasoning module, which supplies hypotheses to the reranker.
The results highlight a clear \textbf{coverage} effect: restricting candidates to a mismatched anomaly family (e.g., \texttt{spike-only}) substantially degrades VUS-PR/AUC-PR, whereas families aligned with the dominant SMD behaviors (e.g., \texttt{shift} or \texttt{seasonal-drift}) are markedly stronger.
This confirms that candidate generation is not a cosmetic component. It determines whether the reranker has access to plausible explanations to directly control final ranking quality.

\subsection{All-types vs budget.}
Enabling \texttt{all types} provides robust performance by avoiding brittle assumptions about anomaly form, but introduces a compute--noise tradeoff governed by the candidate budget (\#Cand).
Increasing the budget improves performance up to an ``elbow'': moving from a low budget (Kcp/maxC=1/1) to the default (3/2) and then to a moderate budget (3/4) steadily increases VUS-PR and yields the best AUC-PR.
Beyond this point, larger budgets (5/4) bring only marginal VUS-PR gains while increasing reranking cost, consistent with diminishing returns due to added distractor candidates.
Therefore, we adopt \textbf{Kcp/maxC=3/4} as the best tradeoff (near-best VUS-PR and best AUC-PR) and use it as our default setting in subsequent experiments.

\paragraph{Why training the reasoner matters (trained vs.\ untrained).} We have justified this and provided detailed justifications in Appendix \ref{sec:appendix:ablation}.

\vspace{-0.25cm}
\section{Conclusion and Limitations}
\label{sec:conclusion}

\paragraph{Conclusion.}
We propose \textsc{RAAS}, a lightweight, detector-agnostic reasoning layer for time-series anomaly detection.
\textsc{RAAS} converts raw detector outputs into event-consistent and explainable scores by (a) extracting compact evidence facts,
(b) reranking a small set of mechanism hypotheses via a YES/NO consistency model, and (c) fusing the induced reasoning score curve with the base detector.
Across univariate and multivariate TSB-AD benchmarks, \textsc{RAAS} consistently improves VUS-PR (and other metrics) for statistical, deep, and zero-shot TSFM detectors, while producing actionable explanations (type, onset, roots, evidence IDs).
\vspace{-0.25cm}
\paragraph{Limitations.}
Our mechanism vocabulary is intentionally small; out-of-vocabulary or multi-mechanism anomalies may be less well supported.
Evidence probes and windowing are hand-designed and may require adjustment under strong non-stationarity or very subtle anomalies.
The reranker is trained on synthetic data; although evidence-space transfer is robust, mismatches between synthetic and real evidence signatures can reduce gains.
Finally, \textsc{RAAS} adds a few inference-time knobs (candidate size, temperature, $\alpha$); automatic tuning and richer temporal consistency constraints are promising future work.

\clearpage

\section*{Impact Statement}
This paper presents a detector-agnostic reasoning layer for time-series anomaly detection, aiming to improve score calibration and provide interpretable explanations (e.g., anomaly type, onset, and responsible variables). The primary positive impact is improved reliability and transparency in monitoring pipelines used in domains such as cloud operations, industrial systems, healthcare, and finance, which can reduce false alarms and enable faster root-cause triage.

Potential risks stem from misuse or over-reliance: (i) incorrect explanations could bias operators toward wrong root causes; (ii) improved detection could be repurposed for intrusive surveillance or monitoring of individuals; and (iii) deployment in high-stakes settings without appropriate validation may lead to harmful decisions. To mitigate these risks, we recommend using the method as a decision-support tool (not a sole decision-maker), reporting uncertainty (e.g., top-$k$ hypotheses and weights), conducting domain-specific validation under distribution shifts, and applying standard privacy/security governance when monitoring sensitive signals.

Overall, we expect the societal implications to align with those of anomaly detection systems broadly—improving safety and efficiency when responsibly deployed—while emphasizing the importance of careful evaluation, human oversight, and context-aware deployment.

\clearpage

\nocite{langley00}

\bibliography{example_paper}
\bibliographystyle{icml2026}

\newpage
\appendix
\onecolumn

\section{Ablation}\label{sec:appendix:ablation}

\begin{figure*}[t]
  \centering
  \begin{subfigure}[t]{0.49\textwidth}
    \centering
    \includegraphics[width=\textwidth]{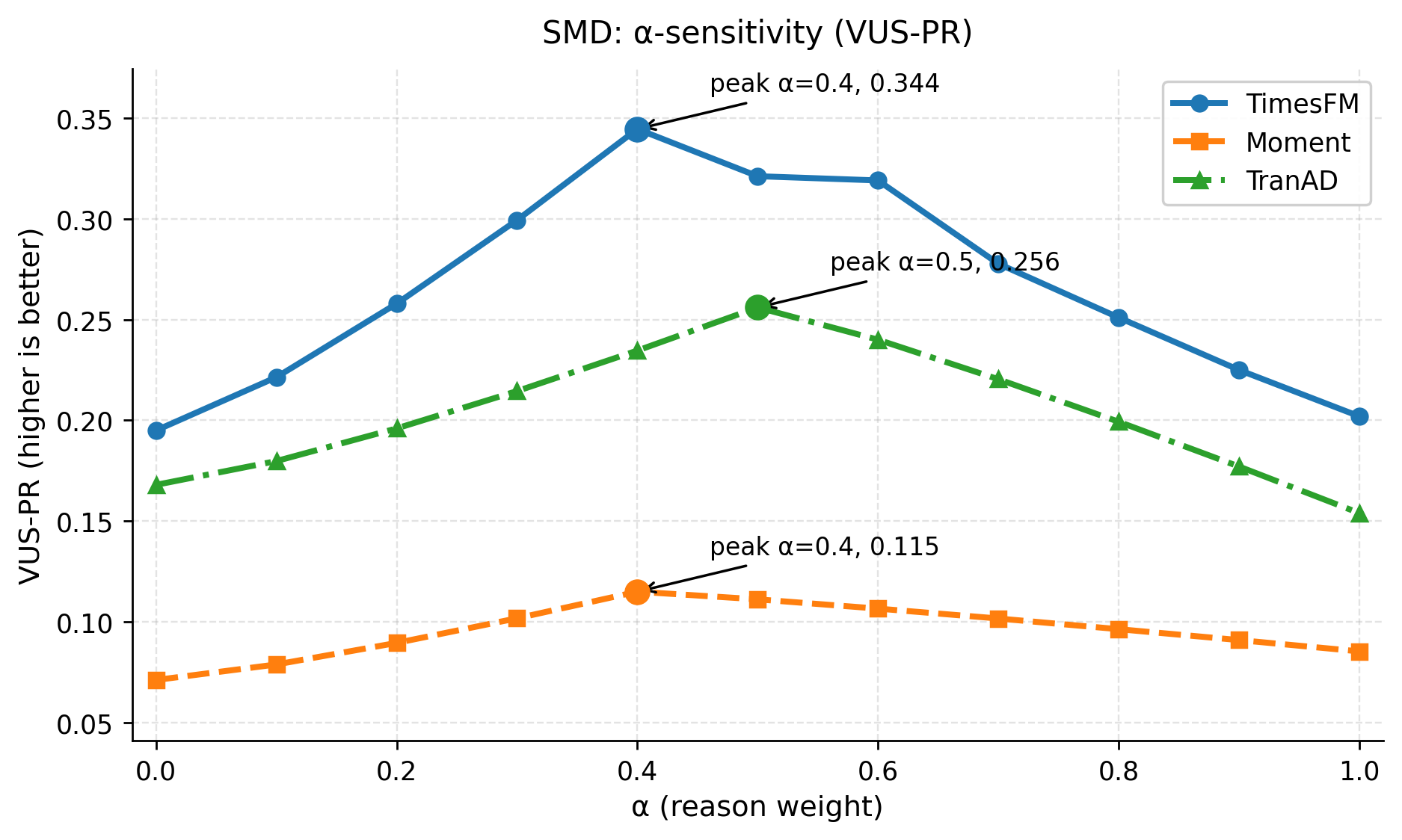}
    \caption{SMD: VUS-PR vs. $\alpha$ (reason weight).}
    \label{fig:alpha_smd_vuspr}
  \end{subfigure}\hfill
  \begin{subfigure}[t]{0.49\textwidth}
    \centering
    \includegraphics[width=\textwidth]{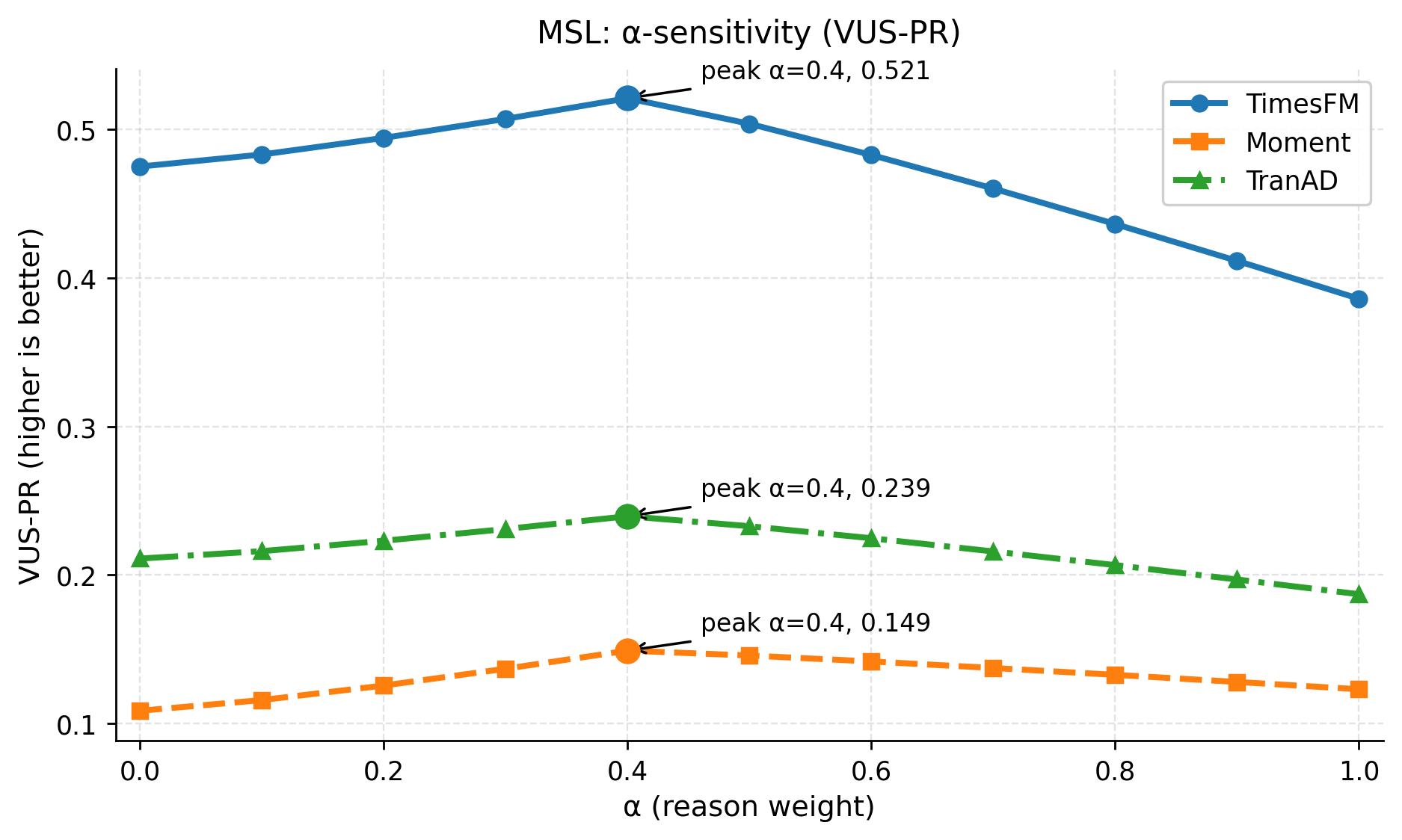}
    \caption{MSL: VUS-PR vs. $\alpha$ (reason weight).}
    \label{fig:alpha_msl_vuspr}
  \end{subfigure}
  \caption{\textbf{Alpha sensitivity of score fusion.}
  We sweep $\alpha\in[0,1]$ in $s_{\alpha}=(1-\alpha)s_{\text{base}}+\alpha s_{\text{reason}}$ and report VUS-PR (higher is better).
  Peaks indicate the best operating point per detector.}
  \label{fig:alpha_sensitivity_vuspr}
\end{figure*}

\begin{figure*}[t]
\vspace{-2mm}
\centering
\setlength{\tabcolsep}{3pt}

\begin{minipage}[t]{0.49\textwidth}
\centering
\includegraphics[width=\linewidth]{images/MSL_IForest_002_MSL_id_1_Sensor_tr_500_1st_900__roots.pdf}
\end{minipage}\hfill
\begin{minipage}[t]{0.49\textwidth}
\centering
\includegraphics[width=\linewidth]{images/MSL_IForest_002_MSL_id_1_Sensor_tr_500_1st_900__scores.pdf}
\end{minipage}

\vspace{1mm}

\begin{minipage}[t]{0.49\textwidth}
\centering
\includegraphics[width=\linewidth]{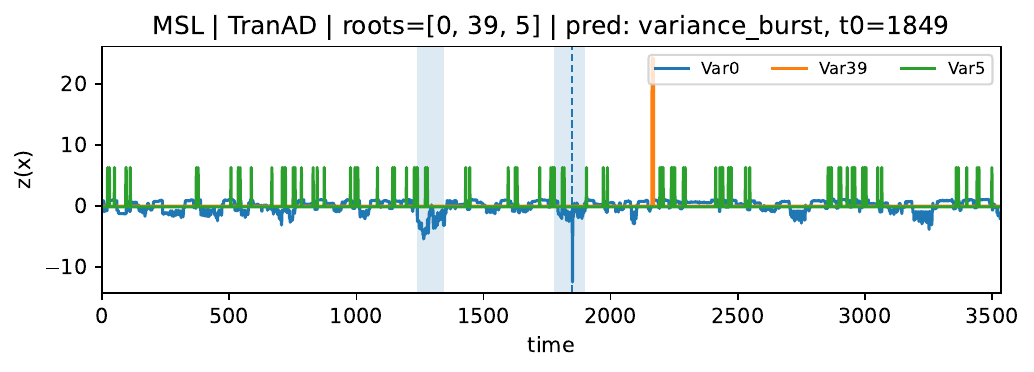}
\end{minipage}\hfill
\begin{minipage}[t]{0.49\textwidth}
\centering
\includegraphics[width=\linewidth]{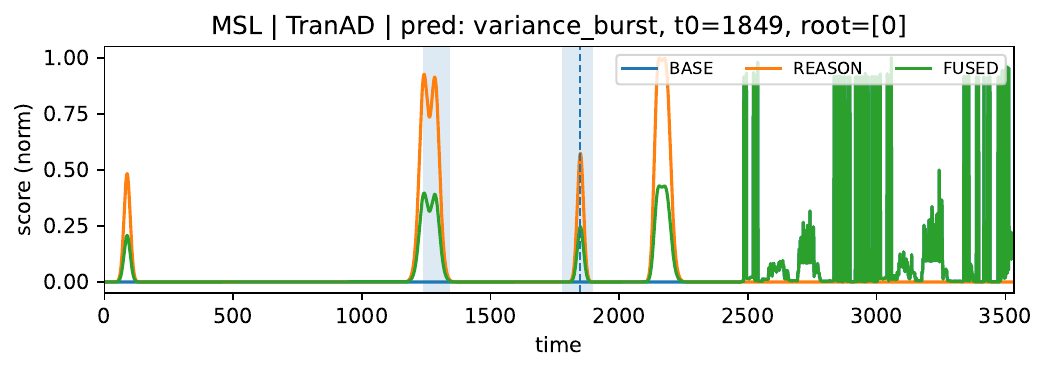}
\end{minipage}

\vspace{1mm}

\begin{minipage}[t]{0.49\textwidth}
\centering
\includegraphics[width=\linewidth]{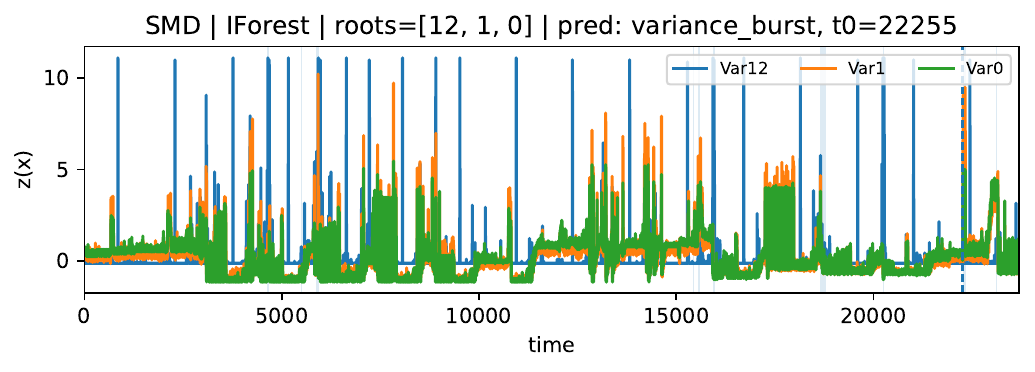}
\end{minipage}\hfill
\begin{minipage}[t]{0.49\textwidth}
\centering
\includegraphics[width=\linewidth]{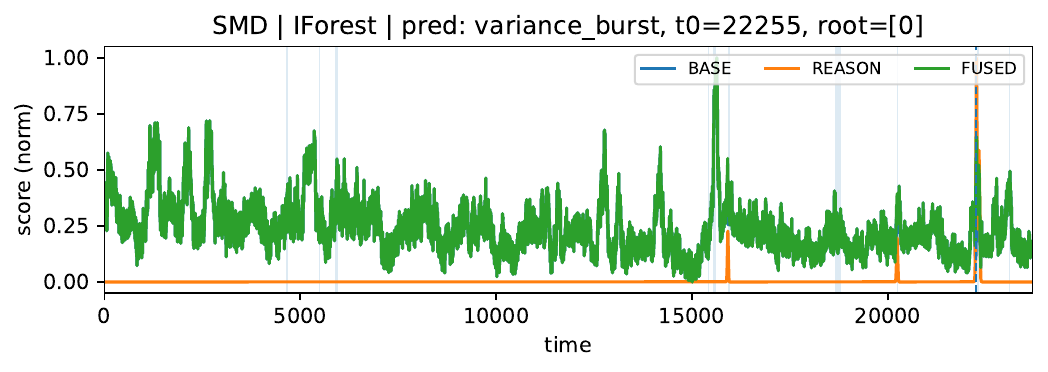}
\end{minipage}

\vspace{1mm}

\begin{minipage}[t]{0.49\textwidth}
\centering
\includegraphics[width=\linewidth]{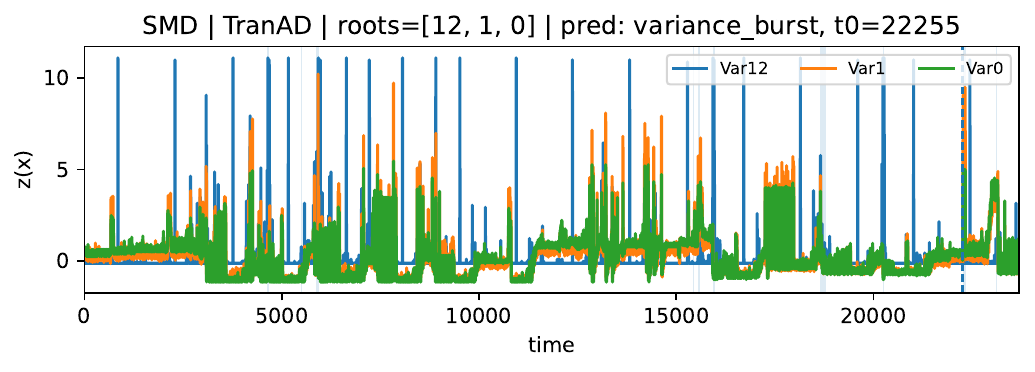}
\end{minipage}\hfill
\begin{minipage}[t]{0.49\textwidth}
\centering
\includegraphics[width=\linewidth]{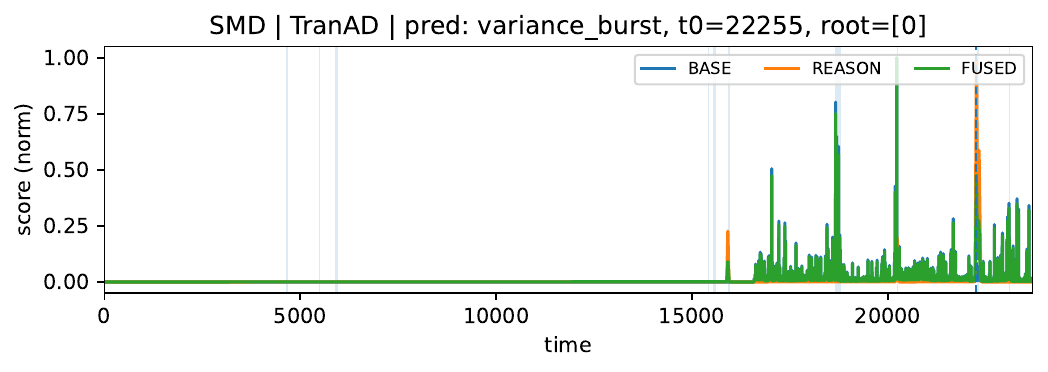}
\end{minipage}

\vspace{-2mm}
\caption{\textbf{Explainability + calibration from RAAS.}
Left panels show \emph{root-variable attribution} and right panels show \emph{score calibration} (Base vs Reason vs Fused).
Across both MSL and SMD cases, the reasoner outputs an event-consistent mechanism with onset $t_0$ and sparse roots, and fusion sharpens localization by concentrating score mass near evidence-supported boundaries.}
\label{fig:explain_cases_8}
\vspace{-3mm}
\end{figure*}

\begin{figure}[t]
    \centering
    \includegraphics[width=0.78\linewidth]{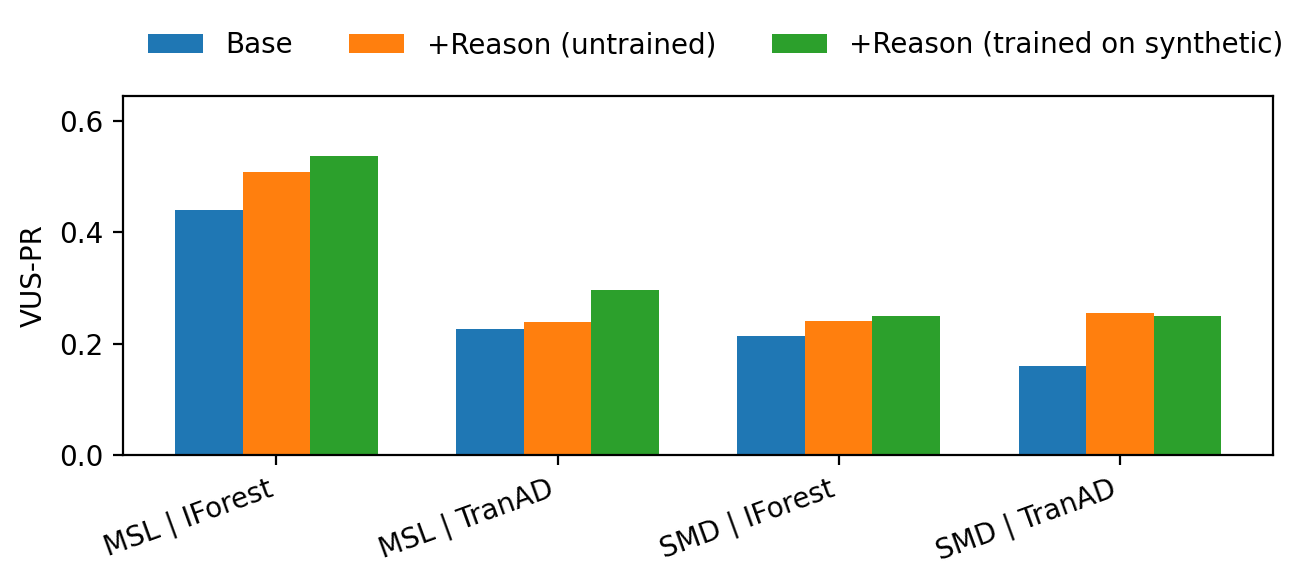}
    \caption{\textbf{Ablation (VUS-PR): untrained vs.\ trained reasoner.}
    Training the reasoner on our synthetic hypothesis--evidence consistency pairs yields consistently higher VUS-PR than using an untrained reasoner, validating that task-specific calibration (YES/NO margins and softmax weighting over hypotheses) is crucial for event-aligned scoring.}
    \label{fig:ablation_untrained_vs_trained_vuspr}
\end{figure}

\paragraph{Why training the reasoner matters (trained vs.\ untrained).}
Our reasoner is used as a \emph{hypothesis--evidence consistency model}: it must assign higher scores to hypotheses whose type/onset/roots match the extracted FACTS and down-rank plausible but incorrect alternatives. An \emph{untrained} LM does not have the task-specific mapping from our structured FACTS format (IDs, numeric probes, peak times, root ranks) to the intended anomaly semantics; as a result, its YES/NO log-probability margins are weakly correlated with true consistency, producing near-uniform or noisy softmax weights over hypotheses. This yields a reasoning score curve that is poorly localized and can even amplify spurious evidence peaks, so fusion provides little benefit.

By contrast, \emph{training on our synthetic consistency pairs} directly optimizes the exact decision used at test time,
\(
\mathrm{Score}_\theta(h)=\log p_\theta(\texttt{YES}\mid \mathcal{F},h)-\log p_\theta(\texttt{NO}\mid \mathcal{F},h),
\)
with hard negatives (wrong type, shifted onset, wrong roots) that force the model to attend to the correct evidence IDs and numeric patterns. This makes the softmax posterior concentrate on a small set of correct hypotheses, sharpening the induced score templates around the true onset and suppressing inconsistent explanations. Consequently, the fused score improves \textbf{VUS-PR} (which rewards correct event/range localization) by reducing broad false-positive regions while preserving recall.

\begin{table}[t]
\centering
\scriptsize
\setlength{\tabcolsep}{3.2pt}
\renewcommand{\arraystretch}{1.12}
\caption{\textbf{Candidate-generation ablation (SMD; TimesFM + Qwen2-0.5B-Instruct; $\alpha=0.3$).}
\#Cand is the average number of candidates per file (reranking cost).
}
\label{tab:candgen_ablation_compact}
\resizebox{0.7\columnwidth}{!}{%
\begin{tabular}{l c c c c}
\toprule
\textbf{Setting} & \textbf{Kcp/maxC} & \textbf{VUS-PR}$\uparrow$ & \textbf{AUC-PR}$\uparrow$ & \textbf{\#Cand}$\downarrow$ \\
\midrule
Type: spike only        & 3/2 & 0.171 & 0.181 & 3  \\
Type: shift only        & 3/2 & 0.265 & 0.248 & 6  \\
Type: seasonal-drift only & 3/2 & 0.285 & 0.264 & 6 \\
\midrule
All types (low budget)  & 1/1 & 0.248 & 0.236 & 15 \\
All types (default)     & 3/2 & 0.264 & 0.244 & 24 \\
All types (best tradeoff) & 3/4 & 0.285 & \textbf{0.265} & 33 \\
All types (best VUS-PR) & 5/4 & \textbf{0.286} & 0.259 & 42 \\
\bottomrule
\end{tabular}}
\end{table}

\paragraph{Qualitative explainability.}
Figure~\ref{fig:explain_cases_8} illustrates that RAAS produces \emph{event-level explanations} in addition to improved scores: it predicts an anomaly mechanism (e.g., \texttt{variance\_burst}) with a concrete onset $t_0$ and a small set of root variables. For example, on MSL the module identifies a variance-burst event with $t_0{=}884$ under IForest and highlights a sparse root set (Var0/Var11/Var49), while the corresponding score plot shows that the fused curve better concentrates around the event instead of spreading mass over broad regions. Similar behavior holds for TranAD on MSL ($t_0{=}1849$) and for both IForest/TranAD on SMD ($t_0{=}22255$), where the fused score aligns more tightly with the evidence-supported event window and reduces spurious activations, yielding more range-consistent detections.

\section{Why Reasoning Improves Anomaly Detection: A Theoretical View}
\label{sec:theory}

Our central claim is that many anomaly detectors produce \emph{high-recall but weakly structured} score sequences, while
real anomalies tend to follow a small set of \emph{mechanistic patterns} (spikes, shifts, variance bursts, trend breaks, missing bursts, etc.).
We formalize our reasoning module as a \emph{mechanism-consistency model} that (i) infers a posterior over a compact hypothesis space from
interpretable evidence, (ii) converts hypotheses into event-shaped score curves, and (iii) fuses this structured signal with any base score.
This provides a principled explanation for why the module improves both ranking metrics (AUC-ROC/PR) and event/range metrics used in TSB-AD.

\paragraph{Setup.}
Let $X\in\mathbb{R}^{T\times D}$ be a multivariate time series and $y_{1:T}\in\{0,1\}^T$ the (unknown) anomaly indicator.
A base detector produces scores $s_b(t)\in\mathbb{R}$, where larger values indicate higher anomaly likelihood.
Let $\Phi(X)$ be a deterministic evidence extractor that outputs a compact set of interpretable facts with IDs,
e.g., change-point peaks, maximal $|\Delta|$, variance jumps, slope breaks, band-energy drift, and missingness bursts.

\paragraph{A mechanism hypothesis space.}
We introduce a structured hypothesis family
\begin{equation}
\mathcal{H}=\{h=(\tau,t_0,\mathcal{R},\mathcal{E})\},
\end{equation}
where $\tau$ is the anomaly type, $t_0$ the onset, $\mathcal{R}\subseteq\{1,\dots,D\}$ a root-variable set,
and $\mathcal{E}$ the subset of evidence IDs used to justify the hypothesis.
Crucially, $\mathcal{H}$ is \emph{small} (a few mechanism types) and admits an efficient candidate subset
$\mathcal{C}(X)\subset\mathcal{H}$ constructed from evidence peaks (top-$K$ change points, max diff, max var jump, etc.),
so inference does not require enumerating $T\times D$ possibilities.

\paragraph{Reasoning as consistency (log-likelihood ratio).}
We train a reasoning model $f_\theta$ to score whether a hypothesis is consistent with the extracted evidence:
\begin{equation}
f_\theta(\Phi(X),h)
\ \approx\
\log\frac{p_\theta(\mathrm{YES}\mid \Phi(X),h)}{p_\theta(\mathrm{NO}\mid \Phi(X),h)}.
\label{eq:lratio_refined}
\end{equation}
In our implementation, $f_\theta$ is computed by a compact causal LM as the margin
$\log p_\theta(\texttt{YES}\mid \Phi(X),h)-\log p_\theta(\texttt{NO}\mid \Phi(X),h)$, which yields a stable discriminative
score with only next-token likelihood evaluation.

\paragraph{Posterior over mechanisms.}
We convert consistency scores into a soft posterior over the candidate set:
\begin{equation}
w(h\mid X)
=
\frac{\exp(f_\theta(\Phi(X),h)/\beta)}{\sum_{h'\in\mathcal{C}(X)}\exp(f_\theta(\Phi(X),h')/\beta)},
\qquad h\in\mathcal{C}(X),
\label{eq:softposterior_refined}
\end{equation}
where $\beta>0$ is a temperature. This posterior admits both (i) a best explanation $\hat h=\arg\max_h f_\theta(\Phi(X),h)$
and (ii) a mixture weight distribution used to construct a smooth score curve.

\paragraph{From mechanisms to event-shaped scores.}
Each hypothesis induces a localized, type-specific score template $k_\tau(\cdot)$:
\begin{equation}
s(t\mid h)
=
k_{\tau}(t;\,t_0,\,\Phi(X),\,\mathcal{R}),
\label{eq:kernel_refined}
\end{equation}
where $k_\tau$ is peaked around $t_0$ and scaled by the evidence magnitude relevant to $\tau$ (e.g., mean-jump for shifts,
max-$|\Delta|$ for spikes, variance jump for variance bursts). The reasoning score is the posterior expectation:
\begin{equation}
s_{\text{reason}}(t)
=
\sum_{h\in\mathcal{C}(X)} w(h\mid X)\, s(t\mid h),
\quad \text{followed by normalization to }[0,1].
\label{eq:reasoncurve_refined}
\end{equation}
This directly enforces \emph{event structure}: rather than arbitrary score fluctuations, $s_{\text{reason}}$ is a mixture of
mechanism-shaped kernels centered at evidence-supported onsets.

\paragraph{Fusion as a Bayes-style correction.}
We fuse the base and reasoning scores via a convex combination:
\begin{equation}
s_{\text{fused}}(t)=\mathrm{Norm}\!\Big((1-\alpha)s_b(t)+\alpha s_{\text{reason}}(t)\Big),
\qquad \alpha\in[0,1].
\label{eq:fusion_refined}
\end{equation}
This is theoretically motivated under a standard log-odds view. Suppose there exist monotone link functions such that
$\ell_b(t)$ and $\ell_r(t)$ approximate log-likelihood ratios from two \emph{complementary} information sources:
\begin{equation}
\ell_b(t)\approx \log\frac{p(y_t=1\mid X)}{p(y_t=0\mid X)},
\qquad
\ell_r(t)\approx \log\frac{p(y_t=1\mid \Phi(X),\mathcal{H})}{p(y_t=0\mid \Phi(X),\mathcal{H})}.
\label{eq:logodds_components}
\end{equation}
If these sources are approximately conditionally independent given $y_t$ (a common ensemble assumption), then the combined
posterior satisfies
\begin{equation}
\log\frac{p(y_t=1\mid X,\Phi(X))}{p(y_t=0\mid X,\Phi(X))}
\ \approx\ \ell_b(t)+\ell_r(t)+\mathrm{const}.
\label{eq:bayes_additive}
\end{equation}
Eq.~\eqref{eq:fusion_refined} is a numerically stable proxy for combining two calibrated signals while preserving ranking
and enabling a tunable trade-off via $\alpha$.

\paragraph{Why this improves event/range metrics (TSB-AD).}
Many detectors produce \emph{diffuse} high-score regions (high recall, low precision) or \emph{fragmented} spikes that split
a single event into multiple alarms. TSB-AD metrics reward temporal coherence (event- and range-level correctness), so these
pathologies hurt Standard-F1/Range-F1 even when AUC is acceptable.
In contrast, Eq.~\eqref{eq:reasoncurve_refined} concentrates mass around evidence-supported onsets and generates coherent,
mechanism-shaped score curves. This acts as an explicit \emph{regularizer on score geometry}: it sharpens boundaries, reduces
spurious peaks, and aligns the score trajectory with event semantics---precisely what range-aware metrics measure.

\paragraph{A variance-reduction view (complementary errors).}
Let $\eta(t)=\mathbb{P}(y_t=1\mid X)$ denote the true anomaly probability. Model the two scores as noisy estimators:
\begin{equation}
s_b(t)=\eta(t)+\epsilon_b(t),
\qquad
s_{\text{reason}}(t)=\eta(t)+\epsilon_r(t),
\end{equation}
with $\mathbb{E}[\epsilon_b]=\mathbb{E}[\epsilon_r]=0$. Since $s_b$ is produced by a detector-specific pipeline while
$s_{\text{reason}}$ is produced by mechanism-consistency on $\Phi(X)$, their error processes are typically weakly correlated.
Then the fused estimator has
\begin{equation}
\mathbb{E}\!\left[(s_{\text{fused}}-\eta)^2\right]
=
(1-\alpha)^2\mathbb{E}[\epsilon_b^2]
+\alpha^2\mathbb{E}[\epsilon_r^2]
+2\alpha(1-\alpha)\mathbb{E}[\epsilon_b\epsilon_r],
\label{eq:mse_refined}
\end{equation}
which is strictly smaller than either component when $\mathbb{E}[\epsilon_b\epsilon_r]$ is not too positive.
This explains detector-agnostic gains: the reasoning module supplies a complementary signal whose mistakes differ from those of
the base detector, improving both ranking and localization after fusion.

\paragraph{Why the reasoner generalizes across datasets and detectors.}
The reasoning module does not memorize dataset-specific patterns. Instead, it operates on $\Phi(X)$, which captures
\emph{mechanism-level invariants} (jumps, spikes, variance changes, slope breaks, spectral drift, missingness) that appear across
domains (servers, industrial sensors, spacecraft telemetry). This encourages invariance-based transfer:
even when amplitudes and noise statistics shift, evidence patterns remain informative, and consistency scoring over a small set
of mechanisms generalizes. As a result, the reasoner behaves like a \emph{structured prior} that can be reused across detectors
and datasets, improving robustness without retraining the base detector.

\paragraph{Practical implication.}
Eqs.~\eqref{eq:softposterior_refined}--\eqref{eq:fusion_refined} show that our method can be viewed as an explainable
mixture-of-mechanisms calibration layer: it reweights evidence-supported hypotheses, produces an event-shaped score curve, and
fuses it with any detector score. This yields both improved detection under TSB-AD metrics and human-readable explanations
(type, onset, roots, evidence IDs) by construction.

\section{Synthetic Completeness and Transferability}
\label{sec:synthetic_completeness}

\paragraph{Why synthetic training is sufficient.}
Our reasoner is \emph{not} trained to detect anomalies from raw signals. It is trained for a narrower, more stable task:
\emph{mechanism-consistency verification}. Given a series $X$ we compute a deterministic evidence summary
$e=\Phi(X)$ (facts with IDs from fixed signal probes), and given a candidate hypothesis $h$ (type, onset, roots)
the reasoner predicts whether $h$ is consistent with $e$.
This shifts learning from domain-specific waveform modeling to \emph{evidence-level logical discrimination},
which depends primarily on whether a hypothesis matches the signatures captured by $\Phi(\cdot)$.

\paragraph{Mechanism vocabulary.}
We fix a finite set of mechanisms
\begin{equation}
\begin{aligned}
\mathcal{T}=\{&\textsc{shift},\textsc{spike},\textsc{variance\_burst},\\
&\textsc{trend\_change},\textsc{seasonal\_drift},\textsc{missing\_burst}\},
\end{aligned}
\end{equation}
and define hypotheses $h=(\tau,t_0,\mathcal{R},\mathcal{E})$ with $\tau\in\mathcal{T}$, onset $t_0$, root set $\mathcal{R}$,
and referenced evidence IDs $\mathcal{E}$. The reasoner is a binary scorer
\begin{equation}
R_\theta(e,h)\in[0,1], \qquad e=\Phi(X),
\end{equation}
implemented as a calibrated YES/NO likelihood margin (Sec.~\ref{sec:method}).

\paragraph{Evidence-level completeness.}
Let $P_{\text{syn}}$ denote the synthetic generator: it samples $(\tau,t_0,\mathcal{R},\xi)$ (mechanism, onset, roots, nuisance)
and produces $X$ by injecting the mechanism into a base signal with nuisance factors $\xi$ (noise level, amplitude, duration, seasonality, correlation).
We say $P_{\text{syn}}$ is \emph{complete for the vocabulary} if it covers, for every $\tau\in\mathcal{T}$:
(i) \emph{positive} pairs where $h$ matches the injected parameters within tolerance, and
(ii) \emph{hard negatives} that are plausible but inconsistent (wrong type, shifted onset, incorrect roots),
such that the induced evidence summaries span the decision boundary the reasoner must learn.

Formally, let $y\in\{0,1\}$ indicate consistency of $(e,h)$.
Completeness means that for each $\tau$ there exists a non-trivial region of evidence space where both labels occur:
\begin{equation}
\forall \tau\in\mathcal{T}\ \exists\ \mathcal{U}_\tau\subseteq \mathrm{range}(\Phi)\ \text{s.t.}\ 
\Pr_{\text{syn}}(e\in\mathcal{U}_\tau, y=1)>0 \ \wedge\  \Pr_{\text{syn}}(e\in\mathcal{U}_\tau, y=0)>0.
\label{eq:completeness_region}
\end{equation}
Intuitively, the generator must expose the reasoner to both correct and confusing alternatives around the same evidence signature.

\paragraph{A weak separability assumption in evidence space.}
The purpose of $\Phi(\cdot)$ is to make mechanism discrimination \emph{easier} than in raw space.
Assume that for each type $\tau$ there exists a type-specific evidence statistic $g_\tau(e)$ (e.g., max-$|\Delta|$ for spikes,
variance jump for variance bursts) such that correct hypotheses achieve larger expected activation near the true onset than incorrect ones.
One simple form is a margin condition: for a true hypothesis $h^\star=(\tau,t_0,\mathcal{R},\cdot)$ and an incorrect $h'$,
\begin{equation}
\mathbb{E}\big[g_\tau(e)\mid (e,h^\star,y{=}1)\big]\ \ge\ \mathbb{E}\big[g_\tau(e)\mid (e,h',y{=}0)\big] + \gamma,
\label{eq:evidence_margin}
\end{equation}
for some $\gamma>0$. This is mild: our probes are explicitly designed so different mechanisms trigger distinct signatures.

\paragraph{Consistency learning implies reliable reranking.}
Training $R_\theta$ on synthetic pairs $(e,h,y)\sim P_{\text{syn}}$ with a proper loss (cross-entropy on the YES/NO token)
yields a calibrated consistency estimator. Under completeness~\eqref{eq:completeness_region} and the evidence-margin condition~\eqref{eq:evidence_margin},
standard generalization arguments imply a low misranking probability on synthetic candidates:
for $h^\star$ consistent and $h'$ inconsistent,
\begin{equation}
\Pr_{\text{syn}}\!\left(R_\theta(e,h^\star) \le R_\theta(e,h')\right)
\ \le\ \varepsilon_{\text{gen}}(n) + \varepsilon_{\text{sep}}(\gamma),
\label{eq:rerank_bound}
\end{equation}
where $\varepsilon_{\text{gen}}(n)$ decreases with training set size and $\varepsilon_{\text{sep}}(\gamma)$ decreases as the evidence margin grows.
This is exactly what we need at inference: selecting top hypotheses from $\mathcal{C}(X)$ and turning them into an event-shaped score curve.

\paragraph{Why synthetic-to-real transfer is easier in evidence space.}
Let $P_{\text{syn}}(e,h)$ and $P_{\text{real}}(e,h)$ denote the induced joint distributions over evidence and hypotheses under
synthetic generation and real data, respectively. For any bounded loss $\ell$, a standard domain shift bound gives
\begin{equation}
\mathcal{R}_{\text{real}}(\theta)
\ \le\ 
\mathcal{R}_{\text{syn}}(\theta)
\ +\ 
\mathrm{TV}\!\left(P_{\text{real}}(e,h),P_{\text{syn}}(e,h)\right),
\label{eq:tv_bound}
\end{equation}
where $\mathrm{TV}$ is total variation distance. The key point is that $\Phi(\cdot)$ intentionally discards nuisance variability
(amplitude scaling, sensor units, background dynamics) while retaining mechanism signatures, so the distribution gap
in evidence space is typically much smaller than in raw space:
\begin{equation}
\mathrm{TV}\!\left(P_{\text{real}}(X),P_{\text{syn}}(X)\right)\ \gg\
\mathrm{TV}\!\left(P_{\text{real}}(\Phi(X)),P_{\text{syn}}(\Phi(X))\right).
\label{eq:evidence_gap}
\end{equation}
Thus, the reasoner can transfer even if synthetic waveforms are not photorealistic: it only needs to learn how evidence patterns
support or contradict mechanism hypotheses.

\paragraph{Why the generator is efficient.}
The label for consistency is \emph{free} under simulation: we know the injected $(\tau,t_0,\mathcal{R})$ and can automatically
construct (i) positives (matched hypotheses) and (ii) hard negatives by perturbing type/onset/roots within controlled tolerances.
This produces balanced supervision targeted at the decision boundaries the reranker must learn, with far less data than would be required
to learn anomaly detection end-to-end.

\paragraph{Connection to detector-agnostic fusion.}
Because the reasoner estimates mechanism consistency in evidence space and converts it to an event-shaped score curve,
it provides complementary information to an arbitrary base detector score. Therefore, fusing
$s_{\text{reason}}$ with $s_b$ (Eq.~\eqref{eq:fusion_refined}) improves localization and precision-based metrics
(AUC-PR/VUS-PR/Standard-F1) without retraining the detector, and explanations $(\tau,t_0,\mathcal{R},\mathcal{E})$ arise as a byproduct.

\section{Detailed results}\label{sec:appendix:detailed-results}

\paragraph{Multivariate results (TSB-AD-M).}
Tables~\ref{tab:adm_aucpr}--\ref{tab:adm_stdf1} demonstrate that integrating our reasoning module improves detection quality across multivariate benchmarks (SMD/MSL/PSM/SAMP) and across detector families.
The largest gains typically appear on \textbf{event- and range-sensitive} measures (e.g., VUS-ROC and Standard-F1), indicating that the reasoner primarily fixes a common failure mode shared by many detectors: \emph{miscalibrated, fragmented score trajectories} that do not align with the onset and extent of true anomaly events.
By extracting structured multivariate evidence and reranking a small set of plausible mechanisms, the module converts scattered peaks into \emph{coherent event-shaped score curves}, improving precision at the event level and reducing false alarms.
Improvements are especially strong for classical statistical baselines, which lack temporal semantics and root attribution, but we also observe consistent gains for deep models, suggesting that even learned temporal representations benefit from post-hoc semantic calibration.
Importantly, \textbf{zero-shot TSFMs} (e.g., TimesFM/MOMENT/Chronos/Time-MoE) also improve without retraining, showing that the reasoning layer provides complementary structure beyond what pretrained models encode, particularly under distribution shifts, missing segments, and cross-sensor interactions.
Overall, these results support our key claim: \emph{reasoning-augmented calibration is detector-agnostic and systematically improves multivariate anomaly scoring while adding interpretable outputs (type/onset/roots).}



\begin{table*}[t]
\centering
\scriptsize
\setlength{\tabcolsep}{1.8pt}
\renewcommand{\arraystretch}{1.02}
\caption{\textbf{AUC-PR comparison with/without the reasoning module} on multivariate anomaly detection (SMD/MSL/PSM).
``Base'' is the detector alone and ``RAAS'' augments it with our reasoning module. $\Delta\%$ is relative improvement.}
\label{tab:adm_aucpr}

\resizebox{\textwidth}{!}{%
\begin{tabular}{l|rrr|rrr|rrr|rrr}
\toprule
\multirow{2}{*}{\textbf{Model}} &
\multicolumn{3}{c|}{\textbf{SMD}} &
\multicolumn{3}{c|}{\textbf{MSL}} &
\multicolumn{3}{c|}{\textbf{PSM}} &
\multicolumn{3}{c}{\textbf{Avg (3 datasets)}} \\
\cmidrule(lr){2-4}\cmidrule(lr){5-7}\cmidrule(lr){8-10}\cmidrule(lr){11-13}
& \textbf{Base} & \textbf{RAAS} & \textbf{$\Delta$\%}
& \textbf{Base} & \textbf{RAAS} & \textbf{$\Delta$\%}
& \textbf{Base} & \textbf{RAAS} & \textbf{$\Delta$\%}
& \textbf{Base} & \textbf{RAAS} & \textbf{$\Delta$\%} \\
\midrule
\multicolumn{13}{l}{\textbf{Statistical (5)}} \\
HBOS    & 0.299 & 0.337 & \pos{+12.7\%} & 0.203 & 0.273 & \pos{+34.1\%} & 0.202 & 0.201 & \negval{-0.3\%} & 0.235 & 0.270 & \pos{+14.9\%} \\
IForest & 0.281 & 0.307 & \pos{+9.3\%}  & 0.374 & 0.449 & \pos{+20.2\%} & 0.203 & 0.203 & \pos{+0.0\%} & 0.286 & 0.320 & \pos{+11.8\%} \\
KNN     & 0.264 & 0.339 & \pos{+28.4\%} & 0.155 & 0.265 & \pos{+70.5\%} & 0.155 & 0.156 & \pos{+0.6\%} & 0.191 & 0.253 & \pos{+32.3\%} \\
LOF     & 0.104 & 0.229 & \pos{+120.8\%}& 0.122 & 0.275 & \pos{+125.9\%}& 0.194 & 0.194 & \pos{+0.0\%} & 0.140 & 0.233 & \pos{+65.9\%} \\
PCA     & 0.483 & 0.502 & \pos{+3.9\%}  & 0.305 & 0.410 & \pos{+34.4\%} & 0.215 & 0.215 & \pos{+0.0\%} & 0.334 & 0.376 & \pos{+12.5\%} \\
\midrule
\multicolumn{13}{l}{\textbf{Deep Learning (5)}} \\
AnomalyTransformer & 0.078 & 0.354 & \pos{+353.7\%} & 0.078 & 0.354 & \pos{+353.8\%} & 0.203 & 0.203 & \pos{+0.0\%} & 0.120 & 0.304 & \pos{+154.1\%} \\
LSTMAD             & 0.224 & 0.255 & \pos{+13.7\%}  & 0.237 & 0.260 & \pos{+9.8\%}   & 0.309 & 0.310 & \pos{+0.4\%} & 0.257 & 0.275 & \pos{+7.1\%} \\
OmniAnomaly         & 0.185 & 0.207 & \pos{+12.3\%}  & 0.206 & 0.237 & \pos{+15.2\%}  & 0.252 & 0.256 & \pos{+1.5\%} & 0.214 & 0.233 & \pos{+8.9\%} \\
TranAD             & 0.199 & 0.250 & \pos{+25.6\%}  & 0.206 & 0.216 & \pos{+4.9\%}   & 0.310 & 0.311 & \pos{+0.4\%} & 0.238 & 0.259 & \pos{+8.8\%} \\
USAD               & 0.210 & 0.237 & \pos{+12.8\%}  & 0.213 & 0.304 & \pos{+42.5\%}  & 0.289 & 0.293 & \pos{+1.4\%} & 0.237 & 0.278 & \pos{+17.3\%} \\
\midrule
\multicolumn{13}{l}{\textbf{Foundation Models (Zero-shot) (from CSV)}} \\
Moment (ZS) & 0.039 & 0.094 & \pos{+142.2\%} & 0.069 & 0.112 & \pos{+62.0\%} & 0.089 & 0.090 & \pos{+0.6\%} & 0.066 & 0.099 & \pos{+49.9\%} \\
TimesFM (ZS)& 0.353 & 0.373 & \pos{+5.6\%}   & 0.353 & 0.373 & \pos{+5.6\%}   & 0.122 & 0.124 & \pos{+1.7\%} & 0.276 & 0.290 & \pos{+5.1\%} \\
\midrule
\textbf{Avg (All models)} &
\textbf{0.217} & \textbf{0.297} & \pos{\textbf{+36.7\%}} &
\textbf{0.210} & \textbf{0.294} & \pos{\textbf{+40.0\%}} &
\textbf{0.232} & \textbf{0.229} & \negval{\textbf{-1.0\%}}  &
\textbf{0.220} & \textbf{0.274} & \pos{\textbf{+24.5\%}} \\
\bottomrule
\end{tabular}}
\end{table*}

\begin{table*}[t]
\centering
\scriptsize
\setlength{\tabcolsep}{1.8pt}
\renewcommand{\arraystretch}{1.02}
\caption{\textbf{AUC-ROC comparison with/without the reasoning module} on multivariate anomaly detection (SMD/MSL/PSM).
``Base'' is the detector alone and ``RAAS'' augments it with our reasoning module. $\Delta\%$ is relative improvement.}
\label{tab:adm_aucroc}

\resizebox{\textwidth}{!}{%
\begin{tabular}{l|rrr|rrr|rrr|rrr}
\toprule
\multirow{2}{*}{\textbf{Model}} &
\multicolumn{3}{c|}{\textbf{SMD}} &
\multicolumn{3}{c|}{\textbf{MSL}} &
\multicolumn{3}{c|}{\textbf{PSM}} &
\multicolumn{3}{c}{\textbf{Avg (3 datasets)}} \\
\cmidrule(lr){2-4}\cmidrule(lr){5-7}\cmidrule(lr){8-10}\cmidrule(lr){11-13}
& \textbf{Base} & \textbf{RAAS} & \textbf{$\Delta$\%}
& \textbf{Base} & \textbf{RAAS} & \textbf{$\Delta$\%}
& \textbf{Base} & \textbf{RAAS} & \textbf{$\Delta$\%}
& \textbf{Base} & \textbf{RAAS} & \textbf{$\Delta$\%} \\
\midrule
\multicolumn{13}{l}{\textbf{Statistical (5)}} \\
HBOS    & 0.753 & 0.779 & \pos{+3.5\%}  & 0.584 & 0.758 & \pos{+30.0\%} & 0.641 & 0.641 & \pos{+0.0\%} & 0.659 & 0.726 & \pos{+10.2\%} \\
IForest & 0.751 & 0.779 & \pos{+3.7\%}  & 0.799 & 0.840 & \pos{+5.1\%}  & 0.646 & 0.646 & \pos{+0.0\%} & 0.732 & 0.755 & \pos{+3.1\%}  \\
KNN     & 0.768 & 0.793 & \pos{+3.3\%}  & 0.592 & 0.800 & \pos{+35.0\%} & 0.609 & 0.610 & \pos{+0.1\%} & 0.656 & 0.734 & \pos{+11.9\%} \\
LOF     & 0.613 & 0.694 & \pos{+13.2\%} & 0.580 & 0.792 & \pos{+36.5\%} & 0.633 & 0.633 & \pos{+0.0\%} & 0.609 & 0.706 & \pos{+15.9\%} \\
PCA     & 0.766 & 0.808 & \pos{+5.4\%}  & 0.773 & 0.821 & \pos{+6.2\%}  & 0.661 & 0.661 & \pos{+0.0\%} & 0.733 & 0.763 & \pos{+4.1\%}  \\
\midrule
\multicolumn{13}{l}{\textbf{Deep Learning (5)}} \\
AnomalyTransformer & 0.528 & 0.843 & \pos{+59.7\%} & 0.518 & 0.804 & \pos{+55.4\%} & 0.650 & 0.650 & \pos{+0.0\%} & 0.565 & 0.766 & \pos{+35.6\%} \\
LSTMAD             & 0.593 & 0.731 & \pos{+23.4\%} & 0.586 & 0.771 & \pos{+31.6\%} & 0.754 & 0.754 & \pos{+0.0\%} & 0.644 & 0.752 & \pos{+16.7\%} \\
OmniAnomaly         & 0.573 & 0.671 & \pos{+17.1\%} & 0.574 & 0.714 & \pos{+24.4\%} & 0.703 & 0.705 & \pos{+0.2\%} & 0.617 & 0.697 & \pos{+13.0\%} \\
TranAD             & 0.588 & 0.747 & \pos{+27.1\%} & 0.585 & 0.729 & \pos{+24.6\%} & 0.754 & 0.754 & \pos{+0.0\%} & 0.642 & 0.743 & \pos{+15.8\%} \\
USAD               & 0.580 & 0.688 & \pos{+18.7\%} & 0.567 & 0.718 & \pos{+26.6\%} & 0.739 & 0.741 & \pos{+0.3\%} & 0.629 & 0.716 & \pos{+13.9\%} \\
\midrule
\multicolumn{13}{l}{\textbf{Foundation Models (Zero-shot) (from CSV)}} \\
Moment (ZS) & 0.504 & 0.577 & \pos{+14.5\%} & 0.475 & 0.676 & \pos{+42.3\%} & 0.377 & 0.377 & \pos{+0.1\%} & 0.452 & 0.544 & \pos{+20.3\%} \\
TimesFM (ZS)& 0.718 & 0.809 & \pos{+12.8\%} & 0.718 & 0.809 & \pos{+12.8\%} & 0.517 & 0.520 & \pos{+0.4\%} & 0.651 & 0.713 & \pos{+9.5\%} \\
\midrule
\textbf{Avg (All models)} &
\textbf{0.654} & \textbf{0.720} & \pos{\textbf{+10.1\%}} &
\textbf{0.612} & \textbf{0.769} & \pos{\textbf{+25.6\%}} &
\textbf{0.615} & \textbf{0.618} & \pos{\textbf{+0.3\%}}  &
\textbf{0.627} & \textbf{0.702} & \pos{\textbf{+12.0\%}} \\
\bottomrule
\end{tabular}}
\end{table*}

\begin{table*}[t]
\centering
\scriptsize
\setlength{\tabcolsep}{1.8pt}
\renewcommand{\arraystretch}{1.02}
\caption{\textbf{VUS-ROC comparison with/without the reasoning module} on multivariate anomaly detection (SMD/MSL/PSM).
``Base'' is the detector alone and ``RAAS'' augments it with our reasoning module. $\Delta\%$ is relative improvement.}
\label{tab:adm_vusroc}

\resizebox{\textwidth}{!}{%
\begin{tabular}{l|rrr|rrr|rrr|rrr}
\toprule
\multirow{2}{*}{\textbf{Model}} &
\multicolumn{3}{c|}{\textbf{SMD}} &
\multicolumn{3}{c|}{\textbf{MSL}} &
\multicolumn{3}{c|}{\textbf{PSM}} &
\multicolumn{3}{c}{\textbf{Avg (3 datasets)}} \\
\cmidrule(lr){2-4}\cmidrule(lr){5-7}\cmidrule(lr){8-10}\cmidrule(lr){11-13}
& \textbf{Base} & \textbf{RAAS} & \textbf{$\Delta$\%}
& \textbf{Base} & \textbf{RAAS} & \textbf{$\Delta$\%}
& \textbf{Base} & \textbf{RAAS} & \textbf{$\Delta$\%}
& \textbf{Base} & \textbf{RAAS} & \textbf{$\Delta$\%} \\
\midrule
\multicolumn{13}{l}{\textbf{Statistical (5)}} \\
HBOS    & 0.812 & 0.831 & \pos{+2.3\%}  & 0.654 & 0.793 & \pos{+21.2\%} & 0.720 & 0.721 & \pos{+0.1\%} & 0.729 & 0.782 & \pos{+7.4\%} \\
IForest & 0.750 & 0.772 & \pos{+2.9\%}  & 0.851 & 0.877 & \pos{+3.1\%}  & 0.723 & 0.723 & \pos{+0.0\%} & 0.775 & 0.791 & \pos{+2.1\%} \\
KNN     & 0.842 & 0.861 & \pos{+2.3\%}  & 0.681 & 0.841 & \pos{+23.5\%} & 0.691 & 0.692 & \pos{+0.1\%} & 0.738 & 0.798 & \pos{+8.2\%} \\
LOF     & 0.730 & 0.797 & \pos{+9.2\%}  & 0.659 & 0.834 & \pos{+26.6\%} & 0.712 & 0.712 & \pos{+0.0\%} & 0.700 & 0.781 & \pos{+11.6\%} \\
PCA     & 0.829 & 0.839 & \pos{+1.2\%}  & 0.805 & 0.847 & \pos{+5.2\%}  & 0.737 & 0.737 & \pos{+0.0\%} & 0.790 & 0.808 & \pos{+2.2\%} \\
\midrule
\multicolumn{13}{l}{\textbf{Deep Learning (5)}} \\
AnomalyTransformer & 0.540 & 0.869 & \pos{+61.0\%} & 0.526 & 0.832 & \pos{+58.2\%} & 0.725 & 0.725 & \pos{+0.0\%} & 0.597 & 0.809 & \pos{+35.6\%} \\
LSTMAD             & 0.593 & 0.729 & \pos{+22.9\%} & 0.584 & 0.790 & \pos{+35.2\%} & 0.820 & 0.820 & \pos{+0.0\%} & 0.666 & 0.780 & \pos{+17.0\%} \\
OmniAnomaly         & 0.574 & 0.669 & \pos{+16.6\%} & 0.578 & 0.727 & \pos{+25.7\%} & 0.774 & 0.776 & \pos{+0.2\%} & 0.642 & 0.724 & \pos{+12.9\%} \\
TranAD             & 0.584 & 0.743 & \pos{+27.2\%} & 0.583 & 0.742 & \pos{+27.2\%} & 0.820 & 0.821 & \pos{+0.1\%} & 0.662 & 0.769 & \pos{+16.2\%} \\
USAD               & 0.579 & 0.686 & \pos{+18.4\%} & 0.572 & 0.732 & \pos{+28.1\%} & 0.808 & 0.809 & \pos{+0.2\%} & 0.653 & 0.742 & \pos{+13.7\%} \\
\midrule
\multicolumn{13}{l}{\textbf{Foundation Models (Zero-shot) (from CSV)}} \\
Moment (ZS) & 0.543 & 0.646 & \pos{+18.8\%} & 0.522 & 0.719 & \pos{+37.8\%} & 0.450 & 0.450 & \pos{+0.0\%} & 0.505 & 0.605 & \pos{+19.8\%} \\
TimesFM (ZS)& 0.793 & 0.810 & \pos{+2.1\%}  & 0.793 & 0.810 & \pos{+2.1\%}  & 0.504 & 0.505 & \pos{+0.3\%} & 0.697 & 0.708 & \pos{+1.5\%} \\
\midrule
\textbf{Avg (All models)} &
\textbf{0.666} & \textbf{0.717} & \pos{\textbf{+7.7\%}}  &
\textbf{0.657} & \textbf{0.798} & \pos{\textbf{+21.3\%}} &
\textbf{0.563} & \textbf{0.566} & \pos{\textbf{+0.7\%}}  &
\textbf{0.629} & \textbf{0.693} & \pos{\textbf{+10.3\%}} \\
\bottomrule
\end{tabular}}
\end{table*}

\begin{table*}[t]
\centering
\scriptsize
\setlength{\tabcolsep}{1.8pt}
\renewcommand{\arraystretch}{1.02}
\caption{\textbf{Standard-F1 comparison with/without the reasoning module} on multivariate anomaly detection (SMD/MSL/PSM).
``Base'' is the detector alone and ``RAAS'' augments it with our reasoning module. $\Delta\%$ is relative improvement.}
\label{tab:adm_stdf1}

\resizebox{\textwidth}{!}{%
\begin{tabular}{l|rrr|rrr|rrr|rrr}
\toprule
\multirow{2}{*}{\textbf{Model}} &
\multicolumn{3}{c|}{\textbf{SMD}} &
\multicolumn{3}{c|}{\textbf{MSL}} &
\multicolumn{3}{c|}{\textbf{PSM}} &
\multicolumn{3}{c}{\textbf{Avg (3 datasets)}} \\
\cmidrule(lr){2-4}\cmidrule(lr){5-7}\cmidrule(lr){8-10}\cmidrule(lr){11-13}
& \textbf{Base} & \textbf{RAAS} & \textbf{$\Delta$\%}
& \textbf{Base} & \textbf{RAAS} & \textbf{$\Delta$\%}
& \textbf{Base} & \textbf{RAAS} & \textbf{$\Delta$\%}
& \textbf{Base} & \textbf{RAAS} & \textbf{$\Delta$\%} \\
\midrule
\multicolumn{13}{l}{\textbf{Statistical (5)}} \\
HBOS    & 0.364 & 0.396 & \pos{+8.8\%}  & 0.266 & 0.373 & \pos{+40.2\%} & 0.246 & 0.246 & \pos{+0.0\%} & 0.292 & 0.338 & \pos{+15.6\%} \\
IForest & 0.351 & 0.378 & \pos{+7.8\%}  & 0.440 & 0.498 & \pos{+13.2\%} & 0.248 & 0.248 & \pos{+0.0\%} & 0.346 & 0.375 & \pos{+8.3\%}  \\
KNN     & 0.323 & 0.384 & \pos{+18.9\%} & 0.236 & 0.356 & \pos{+50.8\%} & 0.235 & 0.236 & \pos{+0.4\%} & 0.265 & 0.325 & \pos{+22.7\%} \\
LOF     & 0.180 & 0.296 & \pos{+64.5\%} & 0.211 & 0.367 & \pos{+73.9\%} & 0.243 & 0.243 & \pos{+0.0\%} & 0.211 & 0.302 & \pos{+43.1\%} \\
PCA     & 0.557 & 0.564 & \pos{+1.2\%}  & 0.404 & 0.490 & \pos{+21.3\%} & 0.259 & 0.259 & \pos{+0.0\%} & 0.407 & 0.437 & \pos{+7.4\%}  \\
\midrule
\multicolumn{13}{l}{\textbf{Deep Learning (5)}} \\
AnomalyTransformer & 0.122 & 0.303 & \pos{+148.6\%} & 0.126 & 0.411 & \pos{+226.4\%} & 0.248 & 0.248 & \pos{+0.0\%} & 0.165 & 0.321 & \pos{+94.5\%} \\
LSTMAD             & 0.296 & 0.340 & \pos{+15.0\%}  & 0.300 & 0.358 & \pos{+19.4\%}  & 0.347 & 0.348 & \pos{+0.3\%} & 0.314 & 0.349 & \pos{+11.2\%} \\
OmniAnomaly         & 0.265 & 0.291 & \pos{+9.7\%}   & 0.271 & 0.319 & \pos{+17.9\%}  & 0.282 & 0.285 & \pos{+1.1\%} & 0.273 & 0.298 & \pos{+9.3\%}  \\
TranAD             & 0.290 & 0.338 & \pos{+16.5\%}  & 0.291 & 0.326 & \pos{+12.0\%}  & 0.348 & 0.348 & \pos{+0.0\%} & 0.310 & 0.337 & \pos{+8.6\%}  \\
USAD               & 0.277 & 0.305 & \pos{+10.1\%}  & 0.273 & 0.373 & \pos{+36.7\%}  & 0.322 & 0.324 & \pos{+0.6\%} & 0.291 & 0.334 & \pos{+14.8\%} \\
\midrule
\multicolumn{13}{l}{\textbf{Foundation Models (Zero-shot) (from CSV)}} \\
Moment (ZS) & 0.102 & 0.187 & \pos{+83.9\%} & 0.093 & 0.168 & \pos{+80.5\%} & 0.201 & 0.201 & \pos{+0.0\%} & 0.132 & 0.185 & \pos{+40.2\%} \\
TimesFM (ZS)& 0.478 & 0.512 & \pos{+7.1\%}  & 0.478 & 0.512 & \pos{+7.1\%}  & 0.228 & 0.230 & \pos{+0.8\%} & 0.395 & 0.418 & \pos{+5.8\%} \\
\midrule
\textbf{Avg (All models)} &
\textbf{0.281} & \textbf{0.362} & \pos{\textbf{+28.9\%}} &
\textbf{0.284} & \textbf{0.376} & \pos{\textbf{+32.6\%}} &
\textbf{0.311} & \textbf{0.311} & \pos{\textbf{+0.0\%}}  &
\textbf{0.292} & \textbf{0.350} & \pos{\textbf{+19.9\%}} \\
\bottomrule
\end{tabular}}
\end{table*}

\section{Detailed results for univariate datasets}\label{sec:appendix:result-univariate}

\paragraph{Univariate results (TSB-AD-U).}
Table~\ref{tab:tsb_adu_aucpr}--\ref{tab:tsb_adu_standardf1} show that adding our reasoning module consistently improves \emph{range-aware} performance across diverse univariate benchmarks.
The gains are most pronounced for lightweight statistical detectors (e.g., HBOS/KNN/LOF), where raw scores are often noisy and poorly aligned with anomaly \emph{events}; the hypothesis-driven reasoner smooths fragmented spikes into coherent event-shaped curves, increasing AUC-PR/VUS-ROC and substantially boosting Standard-F1 by reducing over-triggering and missed ranges.
Deep detectors also benefit, though typically with smaller but steady improvements, indicating that even learned temporal models still exhibit calibration and onset ambiguity that the reasoning layer can correct post hoc.
For zero-shot TSFMs (TimesFM and MOMENT), the reasoner further improves robustness by reconciling model scores with evidence-based hypotheses, yielding consistent average gains across datasets while preserving the plug-and-play nature (no retraining of the base model).
Overall, the results support our claim that \emph{structured reasoning acts as a calibration layer} that upgrades heterogeneous detectors into more event-consistent and interpretable anomaly scoring on univariate data.

\begin{table*}[t]
\centering
\scriptsize
\setlength{\tabcolsep}{1.35pt}
\renewcommand{\arraystretch}{0.98}
\caption{\textbf{Univariate anomaly detection (TSB-AD-U): AUC-PR (higher is better).}
We report \textbf{Base} vs \textbf{RAAS} mean AUC-PR over files for each dataset.
$\Delta\%$ is relative improvement.
\textbf{Avg} summarizes the mean across the 5 datasets, and \textbf{Avg $\Delta\%$} is the relative gain on that mean.}
\label{tab:tsb_adu_aucpr}

\resizebox{\textwidth}{!}{%
\begin{tabular}{ll|rrr|rrr|rrr|rrr|rrr|rrr}
\toprule
\multirow{2}{*}{\textbf{Family}} & \multirow{2}{*}{\textbf{Detector}} &
\multicolumn{3}{c|}{\textbf{IOPS}} &
\multicolumn{3}{c|}{\textbf{MGAB}} &
\multicolumn{3}{c|}{\textbf{NAB}} &
\multicolumn{3}{c|}{\textbf{NEK}} &
\multicolumn{3}{c|}{\textbf{Power}} &
\multicolumn{3}{c}{\textbf{Avg (5 datasets)}} \\
\cmidrule(lr){3-5}\cmidrule(lr){6-8}\cmidrule(lr){9-11}\cmidrule(lr){12-14}\cmidrule(lr){15-17}\cmidrule(lr){18-20}
& &
\textbf{Base} & \textbf{RAAS} & \textbf{$\Delta$\%} &
\textbf{Base} & \textbf{RAAS} & \textbf{$\Delta$\%} &
\textbf{Base} & \textbf{RAAS} & \textbf{$\Delta$\%} &
\textbf{Base} & \textbf{RAAS} & \textbf{$\Delta$\%} &
\textbf{Base} & \textbf{RAAS} & \textbf{$\Delta$\%} &
\textbf{Base} & \textbf{RAAS} & \textbf{$\Delta$\%} \\
\midrule
\multirow{5}{*}{\textbf{Stat}} & IForest & 0.216 & 0.233 & \pos{+7.6\%} & 0.006 & 0.007 & \pos{+16.3\%} & 0.477 & 0.494 & \pos{+3.4\%} & 0.874 & 0.887 & \pos{+1.4\%} & 0.091 & 0.095 & \pos{+4.4\%} & 0.333 & 0.343 & \pos{+2.9\%} \\
 & LOF & 0.085 & 0.131 & \pos{+54.4\%} & 0.006 & 0.008 & \pos{+39.2\%} & 0.326 & 0.454 & \pos{+39.5\%} & 0.628 & 0.868 & \pos{+38.1\%} & 0.101 & 0.109 & \pos{+8.0\%} & 0.229 & 0.314 & \pos{+37.0\%} \\
 & HBOS & 0.014 & 0.196 & \pos{+1326.0\%} & 0.006 & 0.009 & \pos{+48.1\%} & 0.273 & 0.555 & \pos{+103.3\%} & 0.241 & 0.718 & \pos{+197.9\%} & 0.101 & 0.110 & \pos{+9.2\%} & 0.127 & 0.318 & \pos{+149.7\%} \\
 & KNN & 0.014 & 0.196 & \pos{+1326.0\%} & 0.006 & 0.009 & \pos{+48.1\%} & 0.273 & 0.555 & \pos{+103.3\%} & 0.241 & 0.718 & \pos{+197.9\%} & 0.101 & 0.110 & \pos{+9.2\%} & 0.127 & 0.318 & \pos{+149.7\%} \\
 & PCA & 0.255 & 0.270 & \pos{+6.1\%} & 0.007 & 0.008 & \pos{+12.6\%} & 0.549 & 0.558 & \pos{+1.6\%} & 0.915 & 0.919 & \pos{+0.5\%} & 0.093 & 0.095 & \pos{+2.9\%} & 0.364 & 0.370 & \pos{+1.6\%} \\
\midrule
\multirow{5}{*}{\textbf{DL}} & AnomalyTransformer & 0.034 & 0.171 & \pos{+405.0\%} & 0.007 & 0.010 & \pos{+36.4\%} & 0.281 & 0.551 & \pos{+96.1\%} & 0.241 & 0.718 & \pos{+197.9\%} & 0.101 & 0.111 & \pos{+10.2\%} & 0.133 & 0.312 & \pos{+134.7\%} \\
 & TranAD & 0.038 & 0.109 & \pos{+188.0\%} & 0.007 & 0.008 & \pos{+14.0\%} & 0.389 & 0.451 & \pos{+15.8\%} & 0.806 & 0.843 & \pos{+4.7\%} & 0.100 & 0.098 & \negval{-1.8\%} & 0.268 & 0.302 & \pos{+12.8\%} \\
 & USAD & 0.041 & 0.097 & \pos{+138.0\%} & 0.007 & 0.008 & \pos{+11.2\%} & 0.417 & 0.445 & \pos{+6.6\%} & 0.828 & 0.832 & \pos{+0.5\%} & 0.100 & 0.098 & \negval{-2.0\%} & 0.279 & 0.296 & \pos{+6.2\%} \\
 & OmniAnomaly & 0.043 & 0.080 & \pos{+87.3\%} & 0.008 & 0.009 & \pos{+12.1\%} & 0.493 & 0.505 & \pos{+2.5\%} & 0.835 & 0.834 & \negval{-0.1\%} & 0.100 & 0.099 & \negval{-1.5\%} & 0.296 & 0.305 & \pos{+2.9\%} \\
 & LSTMAD & 0.049 & 0.118 & \pos{+142.1\%} & 0.018 & 0.022 & \pos{+21.3\%} & 0.437 & 0.481 & \pos{+10.1\%} & 0.795 & 0.823 & \pos{+3.6\%} & 0.100 & 0.098 & \negval{-2.0\%} & 0.280 & 0.308 & \pos{+10.0\%} \\
\midrule
\multirow{2}{*}{\textbf{FM}} & TimesFM & 0.278 & 0.264 & \negval{-4.9\%} & 0.007 & 0.007 & \pos{+1.7\%} & 0.277 & 0.385 & \pos{+38.9\%} & 0.537 & 0.909 & \pos{+69.4\%} & 0.099 & 0.108 & \pos{+9.4\%} & 0.240 & 0.335 & \pos{+39.9\%} \\
 & MOMENT (ZS) & 0.129 & 0.206 & \pos{+59.4\%} & 0.006 & 0.007 & \pos{+13.1\%} & 0.621 & 0.644 & \pos{+3.7\%} & 0.846 & 0.870 & \pos{+2.9\%} & 0.096 & 0.097 & \pos{+1.0\%} & 0.340 & 0.365 & \pos{+7.4\%} \\
\midrule
\textbf{Avg (All detectors)} &  & \textbf{0.085} & \textbf{0.172} & \textbf{\pos{+101.8\%}} & \textbf{0.008} & \textbf{0.009} & \textbf{\pos{+18.6\%}} & \textbf{0.401} & \textbf{0.506} & \textbf{\pos{+26.2\%}} & \textbf{0.629} & \textbf{0.818} & \textbf{\pos{+30.0\%}} & \textbf{0.099} & \textbf{0.103} & \textbf{\pos{+3.8\%}} & \textbf{0.244} & \textbf{0.322} & \textbf{\pos{+31.8\%}} \\
\bottomrule
\end{tabular}}
\end{table*}

\begin{table*}[t]
\centering
\scriptsize
\setlength{\tabcolsep}{1.35pt}
\renewcommand{\arraystretch}{0.98}
\caption{\textbf{Univariate anomaly detection (TSB-AD-U): AUC-ROC (higher is better).}
We report \textbf{Base} vs \textbf{RAAS} mean AUC-ROC over files for each dataset.
$\Delta\%$ is relative improvement.
\textbf{Avg} summarizes the mean across the 5 datasets, and \textbf{Avg $\Delta\%$} is the relative gain on that mean.}
\label{tab:tsb_adu_aucroc}

\resizebox{\textwidth}{!}{%
\begin{tabular}{ll|rrr|rrr|rrr|rrr|rrr|rrr}
\toprule
\multirow{2}{*}{\textbf{Family}} & \multirow{2}{*}{\textbf{Detector}} &
\multicolumn{3}{c|}{\textbf{IOPS}} &
\multicolumn{3}{c|}{\textbf{MGAB}} &
\multicolumn{3}{c|}{\textbf{NAB}} &
\multicolumn{3}{c|}{\textbf{NEK}} &
\multicolumn{3}{c|}{\textbf{Power}} &
\multicolumn{3}{c}{\textbf{Avg (5 datasets)}} \\
\cmidrule(lr){3-5}\cmidrule(lr){6-8}\cmidrule(lr){9-11}\cmidrule(lr){12-14}\cmidrule(lr){15-17}\cmidrule(lr){18-20}
& &
\textbf{Base} & \textbf{RAAS} & \textbf{$\Delta$\%} &
\textbf{Base} & \textbf{RAAS} & \textbf{$\Delta$\%} &
\textbf{Base} & \textbf{RAAS} & \textbf{$\Delta$\%} &
\textbf{Base} & \textbf{RAAS} & \textbf{$\Delta$\%} &
\textbf{Base} & \textbf{RAAS} & \textbf{$\Delta$\%} &
\textbf{Base} & \textbf{RAAS} & \textbf{$\Delta$\%} \\
\midrule
\multirow{5}{*}{\textbf{Stat}} & IForest & 0.795 & 0.827 & \pos{+4.1\%} & 0.533 & 0.605 & \pos{+13.5\%} & 0.869 & 0.876 & \pos{+0.8\%} & 0.962 & 0.966 & \pos{+0.4\%} & 0.733 & 0.741 & \pos{+1.1\%} & 0.778 & 0.803 & \pos{+3.2\%} \\
 & LOF & 0.585 & 0.670 & \pos{+14.5\%} & 0.545 & 0.623 & \pos{+14.3\%} & 0.758 & 0.837 & \pos{+10.4\%} & 0.896 & 0.956 & \pos{+6.7\%} & 0.731 & 0.745 & \pos{+1.9\%} & 0.703 & 0.766 & \pos{+9.0\%} \\
 & HBOS & 0.500 & 0.660 & \pos{+32.1\%} & 0.545 & 0.617 & \pos{+13.2\%} & 0.711 & 0.816 & \pos{+14.7\%} & 0.820 & 0.945 & \pos{+15.3\%} & 0.731 & 0.744 & \pos{+1.8\%} & 0.662 & 0.756 & \pos{+14.2\%} \\
 & KNN & 0.500 & 0.660 & \pos{+32.1\%} & 0.545 & 0.617 & \pos{+13.2\%} & 0.711 & 0.816 & \pos{+14.7\%} & 0.820 & 0.945 & \pos{+15.3\%} & 0.731 & 0.744 & \pos{+1.8\%} & 0.662 & 0.756 & \pos{+14.2\%} \\
 & PCA & 0.753 & 0.787 & \pos{+4.6\%} & 0.552 & 0.600 & \pos{+8.7\%} & 0.853 & 0.863 & \pos{+1.2\%} & 0.959 & 0.962 & \pos{+0.3\%} & 0.732 & 0.739 & \pos{+1.0\%} & 0.770 & 0.790 & \pos{+2.5\%} \\
\midrule
\multirow{5}{*}{\textbf{DL}} & AnomalyTransformer & 0.507 & 0.668 & \pos{+31.9\%} & 0.552 & 0.622 & \pos{+12.8\%} & 0.744 & 0.841 & \pos{+13.1\%} & 0.820 & 0.945 & \pos{+15.3\%} & 0.731 & 0.744 & \pos{+1.8\%} & 0.671 & 0.764 & \pos{+13.9\%} \\
 & TranAD & 0.541 & 0.621 & \pos{+14.8\%} & 0.552 & 0.594 & \pos{+7.6\%} & 0.838 & 0.843 & \pos{+0.6\%} & 0.942 & 0.950 & \pos{+0.8\%} & 0.731 & 0.727 & \negval{-0.6\%} & 0.721 & 0.747 & \pos{+3.6\%} \\
 & USAD & 0.575 & 0.619 & \pos{+7.6\%} & 0.552 & 0.590 & \pos{+7.0\%} & 0.850 & 0.844 & \negval{-0.7\%} & 0.954 & 0.953 & \negval{-0.1\%} & 0.731 & 0.727 & \negval{-0.6\%} & 0.732 & 0.747 & \pos{+2.1\%} \\
 & OmniAnomaly & 0.600 & 0.606 & \pos{+1.1\%} & 0.552 & 0.587 & \pos{+6.4\%} & 0.863 & 0.860 & \negval{-0.3\%} & 0.957 & 0.954 & \negval{-0.3\%} & 0.731 & 0.728 & \negval{-0.5\%} & 0.741 & 0.747 & \pos{+0.9\%} \\
 & LSTMAD & 0.634 & 0.653 & \pos{+3.0\%} & 0.590 & 0.603 & \pos{+2.2\%} & 0.855 & 0.856 & \pos{+0.1\%} & 0.951 & 0.952 & \pos{+0.1\%} & 0.731 & 0.727 & \negval{-0.6\%} & 0.752 & 0.758 & \pos{+0.8\%} \\
\midrule
\multirow{2}{*}{\textbf{FM}} & TimesFM & 0.786 & 0.759 & \negval{-3.4\%} & 0.563 & 0.564 & \pos{+0.1\%} & 0.821 & 0.874 & \pos{+6.5\%} & 0.891 & 0.978 & \pos{+9.7\%} & 0.731 & 0.741 & \pos{+1.4\%} & 0.758 & 0.783 & \pos{+3.3\%} \\
 & MOMENT (ZS) & 0.747 & 0.796 & \pos{+6.6\%} & 0.552 & 0.564 & \pos{+2.2\%} & 0.939 & 0.944 & \pos{+0.5\%} & 0.960 & 0.961 & \pos{+0.1\%} & 0.745 & 0.751 & \pos{+0.9\%} & 0.789 & 0.803 & \pos{+1.7\%} \\
\midrule
\textbf{Avg (All detectors)} &  & \textbf{0.620} & \textbf{0.693} & \textbf{\pos{+11.7\%}} & \textbf{0.553} & \textbf{0.597} & \textbf{\pos{+7.9\%}} & \textbf{0.823} & \textbf{0.860} & \textbf{\pos{+4.5\%}} & \textbf{0.920} & \textbf{0.955} & \textbf{\pos{+3.9\%}} & \textbf{0.732} & \textbf{0.736} & \textbf{\pos{+0.5\%}} & \textbf{0.730} & \textbf{0.768} & \textbf{\pos{+5.2\%}} \\
\bottomrule
\end{tabular}}
\end{table*}

\begin{table*}[t]
\centering
\scriptsize
\setlength{\tabcolsep}{1.35pt}
\renewcommand{\arraystretch}{0.98}
\caption{\textbf{Univariate anomaly detection (TSB-AD-U): VUS-ROC (higher is better).}
We report \textbf{Base} vs \textbf{RAAS} mean VUS-ROC over files for each dataset.
$\Delta\%$ is relative improvement.
\textbf{Avg} summarizes the mean across the 5 datasets, and \textbf{Avg $\Delta\%$} is the relative gain on that mean.}
\label{tab:tsb_adu_vusroc}

\resizebox{\textwidth}{!}{%
\begin{tabular}{ll|rrr|rrr|rrr|rrr|rrr|rrr}
\toprule
\multirow{2}{*}{\textbf{Family}} & \multirow{2}{*}{\textbf{Detector}} &
\multicolumn{3}{c|}{\textbf{IOPS}} &
\multicolumn{3}{c|}{\textbf{MGAB}} &
\multicolumn{3}{c|}{\textbf{NAB}} &
\multicolumn{3}{c|}{\textbf{NEK}} &
\multicolumn{3}{c|}{\textbf{Power}} &
\multicolumn{3}{c}{\textbf{Avg (5 datasets)}} \\
\cmidrule(lr){3-5}\cmidrule(lr){6-8}\cmidrule(lr){9-11}\cmidrule(lr){12-14}\cmidrule(lr){15-17}\cmidrule(lr){18-20}
& &
\textbf{Base} & \textbf{RAAS} & \textbf{$\Delta$\%} &
\textbf{Base} & \textbf{RAAS} & \textbf{$\Delta$\%} &
\textbf{Base} & \textbf{RAAS} & \textbf{$\Delta$\%} &
\textbf{Base} & \textbf{RAAS} & \textbf{$\Delta$\%} &
\textbf{Base} & \textbf{RAAS} & \textbf{$\Delta$\%} &
\textbf{Base} & \textbf{RAAS} & \textbf{$\Delta$\%} \\
\midrule
\multirow{5}{*}{\textbf{Stat}} & IForest & 0.795 & 0.816 & \pos{+2.7\%} & 0.525 & 0.582 & \pos{+10.8\%} & 0.860 & 0.865 & \pos{+0.6\%} & 0.959 & 0.962 & \pos{+0.3\%} & 0.731 & 0.734 & \pos{+0.4\%} & 0.774 & 0.792 & \pos{+2.4\%} \\
 & LOF & 0.693 & 0.797 & \pos{+15.0\%} & 0.536 & 0.604 & \pos{+12.8\%} & 0.742 & 0.830 & \pos{+11.9\%} & 0.891 & 0.955 & \pos{+7.2\%} & 0.731 & 0.739 & \pos{+1.1\%} & 0.719 & 0.785 & \pos{+9.2\%} \\
 & HBOS & 0.504 & 0.593 & \pos{+17.7\%} & 0.536 & 0.599 & \pos{+11.8\%} & 0.700 & 0.800 & \pos{+14.2\%} & 0.812 & 0.946 & \pos{+16.5\%} & 0.731 & 0.737 & \pos{+0.8\%} & 0.657 & 0.735 & \pos{+11.8\%} \\
 & KNN & 0.504 & 0.593 & \pos{+17.7\%} & 0.536 & 0.599 & \pos{+11.8\%} & 0.700 & 0.800 & \pos{+14.2\%} & 0.812 & 0.946 & \pos{+16.5\%} & 0.731 & 0.737 & \pos{+0.8\%} & 0.657 & 0.735 & \pos{+11.8\%} \\
 & PCA & 0.760 & 0.793 & \pos{+4.4\%} & 0.538 & 0.580 & \pos{+7.9\%} & 0.846 & 0.853 & \pos{+0.8\%} & 0.960 & 0.961 & \pos{+0.2\%} & 0.731 & 0.734 & \pos{+0.4\%} & 0.767 & 0.784 & \pos{+2.2\%} \\
\midrule
\multirow{5}{*}{\textbf{DL}} & AnomalyTransformer & 0.499 & 0.602 & \pos{+20.8\%} & 0.538 & 0.604 & \pos{+12.3\%} & 0.730 & 0.833 & \pos{+14.1\%} & 0.812 & 0.946 & \pos{+16.5\%} & 0.731 & 0.737 & \pos{+0.8\%} & 0.662 & 0.744 & \pos{+12.4\%} \\
 & TranAD & 0.523 & 0.595 & \pos{+13.8\%} & 0.538 & 0.570 & \pos{+6.0\%} & 0.829 & 0.833 & \pos{+0.6\%} & 0.948 & 0.950 & \pos{+0.2\%} & 0.731 & 0.729 & \negval{-0.3\%} & 0.714 & 0.735 & \pos{+3.0\%} \\
 & USAD & 0.556 & 0.595 & \pos{+7.1\%} & 0.538 & 0.568 & \pos{+5.7\%} & 0.842 & 0.833 & \negval{-1.1\%} & 0.954 & 0.953 & \negval{-0.1\%} & 0.731 & 0.729 & \negval{-0.3\%} & 0.724 & 0.736 & \pos{+1.7\%} \\
 & OmniAnomaly & 0.584 & 0.593 & \pos{+1.5\%} & 0.538 & 0.568 & \pos{+5.6\%} & 0.856 & 0.853 & \negval{-0.3\%} & 0.957 & 0.954 & \negval{-0.3\%} & 0.731 & 0.729 & \negval{-0.3\%} & 0.733 & 0.739 & \pos{+0.8\%} \\
 & LSTMAD & 0.612 & 0.623 & \pos{+1.7\%} & 0.567 & 0.571 & \pos{+0.8\%} & 0.847 & 0.846 & \negval{-0.2\%} & 0.952 & 0.952 & \pos{+0.0\%} & 0.731 & 0.729 & \negval{-0.3\%} & 0.742 & 0.744 & \pos{+0.3\%} \\
\midrule
\multirow{2}{*}{\textbf{FM}} & TimesFM & 0.790 & 0.774 & \negval{-2.1\%} & 0.543 & 0.543 & \pos{+0.0\%} & 0.792 & 0.861 & \pos{+8.7\%} & 0.873 & 0.979 & \pos{+12.1\%} & 0.731 & 0.734 & \pos{+0.4\%} & 0.746 & 0.778 & \pos{+4.3\%} \\
 & MOMENT (ZS) & 0.758 & 0.791 & \pos{+4.4\%} & 0.538 & 0.543 & \pos{+0.9\%} & 0.930 & 0.933 & \pos{+0.3\%} & 0.959 & 0.959 & \pos{+0.0\%} & 0.738 & 0.739 & \pos{+0.2\%} & 0.785 & 0.793 & \pos{+1.0\%} \\
\midrule
\textbf{Avg (All detectors)} &  & \textbf{0.632} & \textbf{0.700} & \textbf{\pos{+10.8\%}} & \textbf{0.538} & \textbf{0.580} & \textbf{\pos{+7.7\%}} & \textbf{0.806} & \textbf{0.840} & \textbf{\pos{+4.2\%}} & \textbf{0.919} & \textbf{0.955} & \textbf{\pos{+3.9\%}} & \textbf{0.731} & \textbf{0.734} & \textbf{\pos{+0.4\%}} & \textbf{0.725} & \textbf{0.762} & \textbf{\pos{+5.1\%}} \\
\bottomrule
\end{tabular}}
\end{table*}

\begin{table*}[t]
\centering
\scriptsize
\setlength{\tabcolsep}{1.35pt}
\renewcommand{\arraystretch}{0.98}
\caption{\textbf{Univariate anomaly detection (TSB-AD-U): Standard-F1 (higher is better).}
We report \textbf{Base} vs \textbf{RAAS} mean Standard-F1 over files for each dataset.
$\Delta\%$ is relative improvement.
\textbf{Avg} summarizes the mean across the 5 datasets, and \textbf{Avg $\Delta\%$} is the relative gain on that mean.}
\label{tab:tsb_adu_standardf1}

\resizebox{\textwidth}{!}{%
\begin{tabular}{ll|rrr|rrr|rrr|rrr|rrr|rrr}
\toprule
\multirow{2}{*}{\textbf{Family}} & \multirow{2}{*}{\textbf{Detector}} &
\multicolumn{3}{c|}{\textbf{IOPS}} &
\multicolumn{3}{c|}{\textbf{MGAB}} &
\multicolumn{3}{c|}{\textbf{NAB}} &
\multicolumn{3}{c|}{\textbf{NEK}} &
\multicolumn{3}{c|}{\textbf{Power}} &
\multicolumn{3}{c}{\textbf{Avg (5 datasets)}} \\
\cmidrule(lr){3-5}\cmidrule(lr){6-8}\cmidrule(lr){9-11}\cmidrule(lr){12-14}\cmidrule(lr){15-17}\cmidrule(lr){18-20}
& &
\textbf{Base} & \textbf{RAAS} & \textbf{$\Delta$\%} &
\textbf{Base} & \textbf{RAAS} & \textbf{$\Delta$\%} &
\textbf{Base} & \textbf{RAAS} & \textbf{$\Delta$\%} &
\textbf{Base} & \textbf{RAAS} & \textbf{$\Delta$\%} &
\textbf{Base} & \textbf{RAAS} & \textbf{$\Delta$\%} &
\textbf{Base} & \textbf{RAAS} & \textbf{$\Delta$\%} \\
\midrule
\multirow{5}{*}{\textbf{Stat}} & IForest & 0.267 & 0.281 & \pos{+5.4\%} & 0.030 & 0.032 & \pos{+6.7\%} & 0.384 & 0.391 & \pos{+1.9\%} & 0.924 & 0.927 & \pos{+0.3\%} & 0.095 & 0.101 & \pos{+6.2\%} & 0.340 & 0.346 & \pos{+1.9\%} \\
 & LOF & 0.187 & 0.236 & \pos{+26.1\%} & 0.031 & 0.040 & \pos{+29.8\%} & 0.302 & 0.369 & \pos{+22.0\%} & 0.818 & 0.902 & \pos{+10.2\%} & 0.101 & 0.114 & \pos{+13.1\%} & 0.288 & 0.332 & \pos{+15.1\%} \\
 & HBOS & 0.027 & 0.286 & \pos{+968.0\%} & 0.031 & 0.040 & \pos{+29.8\%} & 0.282 & 0.542 & \pos{+92.2\%} & 0.654 & 0.923 & \pos{+41.1\%} & 0.101 & 0.114 & \pos{+13.1\%} & 0.219 & 0.381 & \pos{+73.9\%} \\
 & KNN & 0.027 & 0.286 & \pos{+968.0\%} & 0.031 & 0.040 & \pos{+29.8\%} & 0.282 & 0.542 & \pos{+92.2\%} & 0.654 & 0.923 & \pos{+41.1\%} & 0.101 & 0.114 & \pos{+13.1\%} & 0.219 & 0.381 & \pos{+73.9\%} \\
 & PCA & 0.342 & 0.353 & \pos{+3.2\%} & 0.039 & 0.044 & \pos{+13.4\%} & 0.429 & 0.436 & \pos{+1.6\%} & 0.936 & 0.937 & \pos{+0.1\%} & 0.095 & 0.101 & \pos{+6.2\%} & 0.368 & 0.374 & \pos{+1.6\%} \\
\midrule
\multirow{5}{*}{\textbf{DL}} & AnomalyTransformer & 0.052 & 0.284 & \pos{+443.0\%} & 0.031 & 0.040 & \pos{+29.8\%} & 0.296 & 0.545 & \pos{+84.1\%} & 0.654 & 0.923 & \pos{+41.1\%} & 0.101 & 0.115 & \pos{+14.0\%} & 0.227 & 0.381 & \pos{+68.0\%} \\
 & TranAD & 0.110 & 0.284 & \pos{+158.7\%} & 0.039 & 0.044 & \pos{+13.4\%} & 0.372 & 0.391 & \pos{+5.2\%} & 0.910 & 0.923 & \pos{+1.4\%} & 0.101 & 0.099 & \negval{-1.6\%} & 0.306 & 0.348 & \pos{+13.6\%} \\
 & USAD & 0.115 & 0.268 & \pos{+133.0\%} & 0.039 & 0.044 & \pos{+13.4\%} & 0.388 & 0.391 & \pos{+0.8\%} & 0.923 & 0.923 & \pos{+0.0\%} & 0.101 & 0.099 & \negval{-1.6\%} & 0.313 & 0.345 & \pos{+10.2\%} \\
 & OmniAnomaly & 0.123 & 0.286 & \pos{+133.3\%} & 0.044 & 0.046 & \pos{+5.5\%} & 0.409 & 0.416 & \pos{+1.6\%} & 0.925 & 0.923 & \negval{-0.2\%} & 0.101 & 0.099 & \negval{-1.6\%} & 0.320 & 0.354 & \pos{+10.4\%} \\
 & LSTMAD & 0.155 & 0.282 & \pos{+81.7\%} & 0.100 & 0.104 & \pos{+4.1\%} & 0.396 & 0.402 & \pos{+1.4\%} & 0.923 & 0.923 & \pos{+0.0\%} & 0.101 & 0.099 & \negval{-1.6\%} & 0.335 & 0.362 & \pos{+7.9\%} \\
\midrule
\multirow{2}{*}{\textbf{FM}} & TimesFM & 0.343 & 0.333 & \negval{-3.0\%} & 0.040 & 0.040 & \pos{+0.0\%} & 0.351 & 0.440 & \pos{+25.4\%} & 0.833 & 0.967 & \pos{+16.0\%} & 0.099 & 0.114 & \pos{+15.3\%} & 0.333 & 0.379 & \pos{+13.9\%} \\
 & MOMENT (ZS) & 0.212 & 0.282 & \pos{+33.1\%} & 0.040 & 0.040 & \pos{+0.0\%} & 0.545 & 0.545 & \pos{+0.0\%} & 0.936 & 0.936 & \pos{+0.0\%} & 0.101 & 0.104 & \pos{+2.9\%} & 0.367 & 0.381 & \pos{+3.9\%} \\
\midrule
\textbf{Avg (All detectors)} &  & \textbf{0.163} & \textbf{0.291} & \textbf{\pos{+78.6\%}} & \textbf{0.045} & \textbf{0.049} & \textbf{\pos{+9.1\%}} & \textbf{0.378} & \textbf{0.442} & \textbf{\pos{+16.9\%}} & \textbf{0.847} & \textbf{0.930} & \textbf{\pos{+9.8\%}} & \textbf{0.100} & \textbf{0.108} & \textbf{\pos{+7.9\%}} & \textbf{0.307} & \textbf{0.364} & \textbf{\pos{+18.7\%}} \\
\bottomrule
\end{tabular}}
\end{table*}


\begin{figure*}[t]
    \centering
    \includegraphics[width=\textwidth]{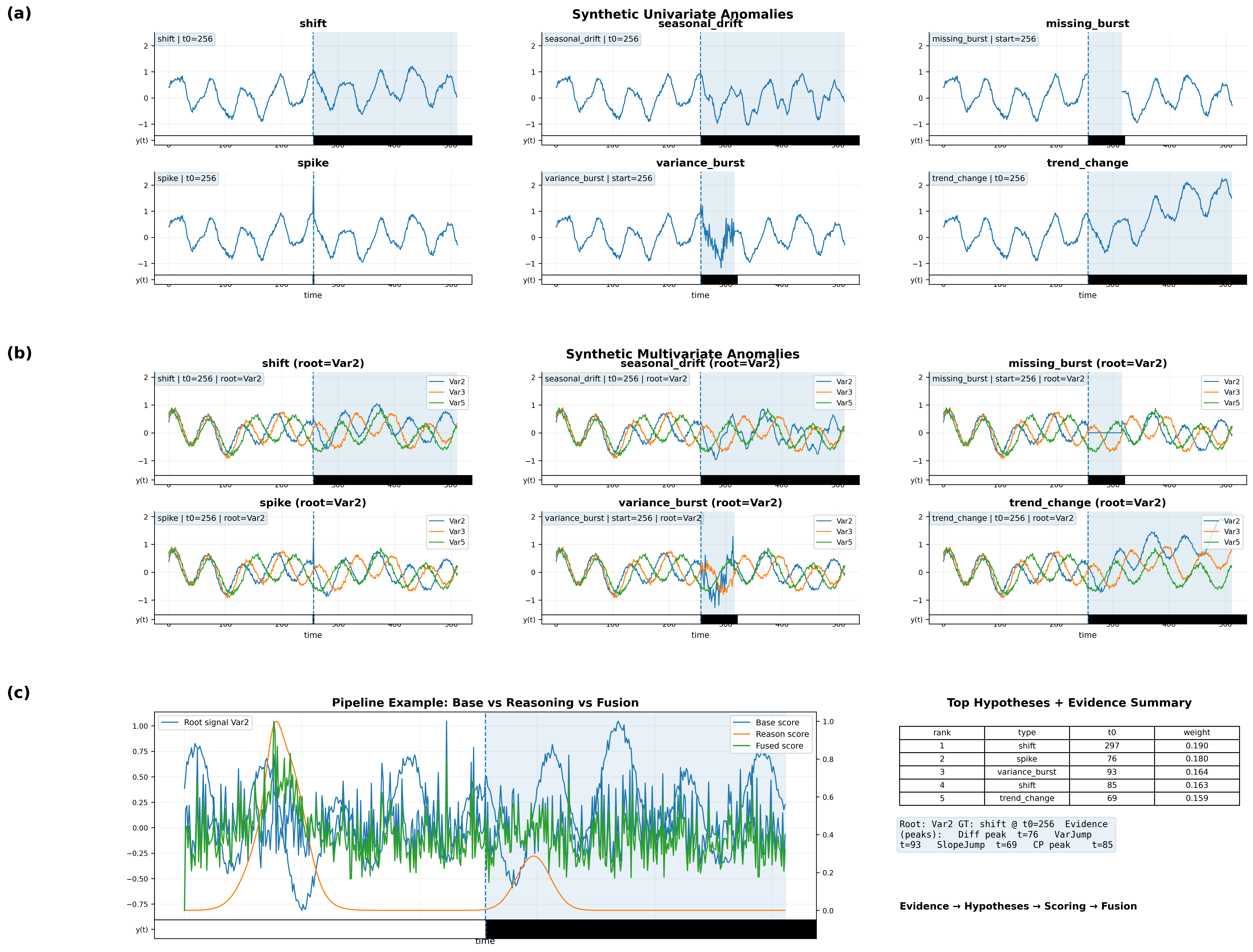}
    \caption{
    \textbf{Reasoning-guided anomaly detection pipeline (synthetic + full flow).}
    (a) Synthetic univariate anomaly patterns used to train the reasoning module, shown with a label-strip (red) and the anomaly start time $t_0$.
    (b) Synthetic multivariate anomalies injected into a root variable (and propagated to related variables), again with ground-truth label-strip and $t_0$.
    (c) End-to-end inference example: baseline detector score, reasoning score produced from evidence-driven hypothesis reranking, and final fused score.
    }
    \label{fig:reasoning_abc}
\end{figure*}


%

\subsection{Synthetic Training Data for the Reranker}\label{subsec:appendix:synthetic}
We train $\mathcal{R}_\theta$ entirely on synthetic time series with controllable anomaly mechanisms,
so it generalizes across detectors and datasets.

\paragraph{Normal series generator.}
We sample a clean signal $x(t)$ from a mixture of components:
\begin{equation}
x(t) = \sum_{k=1}^{K_s} A_k \sin(2\pi f_k t + \phi_k) + \sum_{j=1}^{K_a} \mathrm{AR}_j(t) + \epsilon(t),
\end{equation}
where $\epsilon(t)\sim\mathcal{N}(0,\sigma^2)$.
For multivariate series, we add cross-variable correlations:
\begin{equation}
X(t) = B z(t) + \eta(t),
\end{equation}
with latent sources $z(t)$ and mixing matrix $B$.

\paragraph{Anomaly injection.}
We inject one anomaly with parameters $(\tau,t_0,\mathcal{R})$:
\begin{itemize}[leftmargin=*]
\item \textsc{shift}: $X_{t\ge t_0,d} \leftarrow X_{t,d} + \delta_d$ for $d\in\mathcal{R}$.
\item \textsc{spike}: add impulse(s) at $t_0$ for $d\in\mathcal{R}$.
\item \textsc{variance\_burst}: amplify noise $\epsilon(t)$ on an interval around $t_0$.
\item \textsc{trend\_change}: change slope after $t_0$.
\item \textsc{seasonal\_drift}: change frequency/amplitude after $t_0$.
\item \textsc{missing\_burst}: replace a segment with NaNs or constant fill.
\end{itemize}
The ground truth label sequence $y(t)$ is 1 on the injected anomaly window and 0 otherwise.

\paragraph{Training pairs.}
For each synthetic series, we compute facts $\mathcal{F}$ and sample hypotheses $h$.
We assign a binary label $y_h\in\{0,1\}$ by
\begin{equation}
y_h \;=\; \mathbb{I}\!\left[\,h \approx (\tau,t_0,\mathcal{R})\,\right],
\end{equation}
where $\approx$ denotes a match within preset tolerances. All mismatches (wrong type,
wrong onset, wrong roots) and hypotheses that contradict the evidence are treated as
negatives. We train $\mathcal{R}_\theta$ to output \textsc{yes} for $y_h{=}1$ and
\textsc{no} for $y_h{=}0$. We balance positives/negatives (e.g., $1\!:\!3$) and
randomize fact ordering for robustness.

\paragraph{Objective.}
We fine-tune the LM via next-token prediction on the final answer token:
\begin{equation}
\min_{\theta}\; -\mathbb{E}_{(\mathcal{F},h,y_h)}
\left[
\begin{aligned}
&y_h\,\log p_{\theta}(\textsc{yes}\mid \mathcal{F},h) \\
&+(1-y_h)\,\log p_{\theta}(\textsc{no}\mid \mathcal{F},h)
\end{aligned}
\right].
\end{equation}

\begin{table*}[t]
\centering
\small
\setlength{\tabcolsep}{4.2pt}
\renewcommand{\arraystretch}{1.15}
\caption{\textbf{Reasoning-only anomaly scoring performance on multivariate TSB-AD datasets.}
We evaluate our reasoning module alone (without any base detector) using the official TSB\_AD metrics.
Higher is better for all metrics.}
\label{tab:reasoning_only_multivar_llms}
\resizebox{\textwidth}{!}{
\begin{tabular}{lccccccccc|ccccccccc}
\toprule
\multirow{2}{*}{\textbf{Dataset}} &
\multicolumn{9}{c|}{\textbf{Lightweight Reasoner (GPT-Neo-125M checkpoint)}} &
\multicolumn{9}{c}{\textbf{Qwen2-0.5B-Instruct}} \\
\cmidrule(lr){2-10}\cmidrule(lr){11-19}
&
\textbf{AUC-ROC} &
\textbf{AUC-PR} &
\textbf{VUS-ROC} &
\textbf{VUS-PR} &
\textbf{Std-F1} &
\textbf{PA-F1} &
\textbf{Event-F1} &
\textbf{R-F1} &
\textbf{Aff-F} &
\textbf{AUC-ROC} &
\textbf{AUC-PR} &
\textbf{VUS-ROC} &
\textbf{VUS-PR} &
\textbf{Std-F1} &
\textbf{PA-F1} &
\textbf{Event-F1} &
\textbf{R-F1} &
\textbf{Aff-F} \\
\midrule
SMD  &
0.70896 & 0.32474 & 0.61450 & 0.23403 & 0.40402 & 0.58830 & 0.42130 & 0.34626 & 0.54100 &
0.71478 & 0.31186 & 0.61290 & 0.23483 & 0.41292 & 0.57650 & 0.38810 & 0.32652 & 0.53053 \\

PSM  &
0.51313 & 0.12434 & 0.50131 & 0.12371 & 0.20149 & 0.34974 & 0.07551 & 0.05744 & 0.08670 &
0.51063 & 0.11794 & 0.49905 & 0.12347 & 0.20149 & 0.34878 & 0.07213 & 0.05636 & 0.08701 \\

MSL  &
0.82639 & 0.30453 & 0.84365 & 0.33370 & 0.42309 & 0.75031 & 0.52127 & 0.45028 & 0.80521 &
0.86831 & 0.45001 & 0.88104 & 0.50112 & 0.49370 & 0.84289 & 0.71180 & 0.49549 & 0.92454 \\

SMAP &
0.80424 & 0.21801 & 0.80329 & 0.24059 & 0.32113 & 0.62721 & 0.39002 & 0.35661 & 0.68028 &
0.87118 & 0.33569 & 0.88005 & 0.37070 & 0.42042 & 0.80611 & 0.67507 & 0.48134 & 0.83286 \\

\bottomrule
\end{tabular}
}
\end{table*}

\begin{algorithm}[t]
\caption{Detector-agnostic reasoning-augmented anomaly scoring}
\label{alg:reason_ad}
\begin{algorithmic}[1]
\REQUIRE Time series $X$, base detector $\mathcal{A}$, reasoning LM $\mathcal{R}_\theta$, fusion weight $\alpha$
\STATE $s_{\text{base}}(t) \leftarrow \mathcal{A}(X)_t$
\STATE $\mathcal{F} \leftarrow \mathcal{E}(X)$ \COMMENT{evidence facts with IDs}
\STATE $\mathcal{H} \leftarrow \mathcal{G}(\mathcal{F})$ \COMMENT{candidate hypotheses}
\FOR{each $h_m \in \mathcal{H}$}
    \STATE $\mathrm{Score}_\theta(h_m) \leftarrow \log p_\theta(\texttt{YES}\mid \mathcal{F},h_m) - \log p_\theta(\texttt{NO}\mid \mathcal{F},h_m)$
\ENDFOR
\STATE select top $M'$ hypotheses and compute weights $\{w_m\}$ via softmax
\STATE $s_{\text{reason}}(t) \leftarrow \sum_m w_m \cdot s_{\tau_m}(t; t_{0,m})$
\STATE $s_{\text{fused}}(t) \leftarrow \mathrm{Norm}((1-\alpha)s_{\text{base}}(t) + \alpha s_{\text{reason}}(t))$
\STATE $\hat{h}\leftarrow \arg\max_{h_m}\mathrm{Score}_\theta(h_m)$
\STATE \textbf{Return} $s_{\text{fused}}(t)$ and explanation $\hat{h}$
\end{algorithmic}
\end{algorithm}

\end{document}